\documentclass{article}

\usepackage[dblblindworkshop, final]{neurips_2026}
\workshoptitle{Sim2Science: ML with Imperfect Scientific Models}

\usepackage[utf8]{inputenc}
\usepackage[T1]{fontenc}
\usepackage{hyperref}
\usepackage{url}
\usepackage{booktabs}
\usepackage{amsfonts}
\usepackage{amsmath}
\usepackage{nicefrac}
\usepackage{microtype}
\usepackage{graphicx}
\usepackage{xcolor}
\usepackage{caption}
\usepackage{subcaption}

\newcommand{\reg}{\mathrm{Reg}}

\title{Fidelity Regret: Measuring Simulator Misspecification in Verifiable Design}

\author{%
  Noah Flynn \\
  University of California, Berkeley \\
  \texttt{noahflynn@berkeley.edu} \\
}

\begin{document}

\maketitle

\begin{abstract}
Reinforcement learning from verifiable feedback assumes that the verifier evaluates the
quantity of interest. In genetic-circuit design, a topological
verifier can establish that a program is a valid circuit computing the target function,
whereas ranking its physical performance requires a simulator. We measure \emph{fidelity
regret}, the reference simple regret incurred when a cheap simulator selects the design,
exactly over a 21{,}362-design combination subspace spanning 5 measured gate
libraries and 3 organisms. Exact ordered-assignment checks retain positive regret in every
tested cell while shifting magnitudes in both directions.  Mean normalized regret is 0.369 over
72 cells; 25 cells
(34.7\%) have regret at least 0.5, and only 6 have zero regret. The discrepancy depends
strongly on the reference axis: the yeast library \texttt{SC1C1G1T1} has mean regret 0.113
for cytometry quality and 0.801 for growth burden, and every evaluated library--circuit-size
pair has positive burden regret. Topological verification ties all valid gate assignments,
while an oracle whitelist subject to retaining at least 25\% of the design space leaves
positive regret in 30 of 52 cells. Design-space optimization and feedback-guided search with frozen
language models reach the high-regret regions; on a matched zero-regret control, additional
optimization instead reduces regret. At a matched modeled-cost budget, mitigation is budget- and library-dependent. An in-sample fixed
rule spending $\alpha{=}0.34$ of the budget re-scoring top cheap-simulator designs achieves
mean excess regret 0.016, compared with 0.305 for cheap-only and 0.034 for reference-only selection,
improving on cheap-only in 330 of 360 cells. More broadly, any workflow optimized through a
simulator or digital twin can satisfy its verifier yet perform poorly on the intended
physical objective. Fidelity regret provides a transferable template for detecting this
gap before optimization and comparing budget-aware remedies across scientific domains.
\end{abstract}

\section{Introduction}

Reinforcement learning with verifiable rewards (RLVR) replaces a learned reward with a
checker that directly evaluates task completion \citep{lambert2024}. This connection between reward and task
objective holds in settings such as theorem proving and unit-tested code. In scientific
design, a checker may certify only a necessary condition, leaving a simulator to evaluate
the physical quantity being optimized.

Genetic-circuit design exposes this separation. Using \texttt{pysbol3}
\citep{mitchell2022}, a verifier can validate source syntax, SBOL structure, requested topology,
and target truth table. GenCircuit-RL \citep{gencircuitrl} trains language models with a
multiplicative five-level verifier; its topology tasks use symbolic truth-table propagation, while its
separate T8 assignment evaluation adds Hill-response simulation. Here, topology-only
verification gives every valid measured-gate assignment the same score, despite cell-dependent
physical performance. Simulators must therefore rank valid designs, at the cost of repeated
evaluations during optimization.

Before optimization, we measure fixed reward misspecification exactly over enumerable cells.
Fidelity regret is the reference simple regret of the cheap simulator's argmax; enumerating each
(library, topology, cheap/reference-simulator pair) links it to multi-fidelity screening regret
\citep{pan2022,lee2026}. Across 5 Cello UCF libraries, regret varies by axis; an oracle bound
shows the evaluated per-gate-admissibility verifier class cannot eliminate it across all cells.
Design-space and frozen-model optimizers reach high-regret regions; 1{,}280 matched-budget runs
show mitigation depends on reference axis, cost, and library, motivating fixed-budget re-scoring.

\section{Fidelity ladder and fidelity regret}
\label{sec:def}

\textbf{The ladder.} Five simulator levels score the same design using parameters from the
measured Cello UCF tables \citep{nielsen2016,jones2022}: \textbf{L0}, topology only; \textbf{L1},
mean-field Hill steady state; \textbf{L2}, single-cell cytometry quality (with
\textbf{L2\,$d'$} for separation); \textbf{L3}, growth burden; and \textbf{L4}, response
kinetics. The in-optimization cost model used for every budget comparison prices one L2
evaluation at 67.4 L1 evaluations and one L3 evaluation at 2.2.
Appendix~\ref{app:ladder} reports the separate standalone timings and reconciles the two
measurements.

The levels score distinct physical properties, so a gap combines a different objective with
a more faithful model of the same objective. The falsification control in Section~\ref{sec:opt}
isolates the latter. L2 $d'$ measures ON/OFF separation irrespective of polarity
(Appendix~\ref{app:laddereq}).

\textbf{The quantity.} For a design space $\mathcal{D}$, let
$d_{\text{cheap}} \in \arg\max_{d \in \mathcal{D}}s_{\text{cheap}}(d)$ denote the cheap
simulator's selected optimum. Fidelity regret is
\begin{equation}
\reg = \max_{d \in \mathcal{D}} s_{\text{ref}}(d) - s_{\text{ref}}(d_{\text{cheap}}),
\label{eq:regret}
\end{equation}
which measures the reference quality forfeited by the cheap selection. The reported value
uses the deterministic argmax returned by the enumeration implementation, and the tables
retain the number of tied optima in every cell. Reported regret is conditional on this
recorded selection when L1 has multiple optima; the fully tied L0 case is treated separately
below. We
normalize regret by the reference range over $\mathcal{D}$, so 0 identifies a reference
optimum and 1 the worst enumerated design. We also report the percentile and absolute value
because normalization can magnify narrow reference ranges: \texttt{Bth1C1G1T1}'s worst L4
cell is 0.870 normalized and 0.085 decades absolute. The primary enumeration covers a
combination subspace: an unordered choice of $k$ distinct gates evaluated once in the
implementation's fixed slot order. It contains 21{,}362 designs; the corresponding
ordered-assignment space has $k!$ times as many designs in each $k$-specific cell and totals
432{,}372 designs. Exact
ordered-assignment checks retain positive regret in all eight tested cells, but shift
magnitudes in both directions (mean absolute change 0.273). The reported statistics are exact
for the combination subspace and indicative, not conservative, for the ordered space
(Appendix~\ref{app:ordered}). The reference composes per-gate wet-lab measurements through the
fidelity ladder; it is not a whole-circuit wet-lab measurement (\S\ref{sec:limits}).

\section{Axis-dependent fidelity regret}
\label{sec:enum}

\begin{figure}[t]
  \centering
  \includegraphics[width=0.68\textwidth]{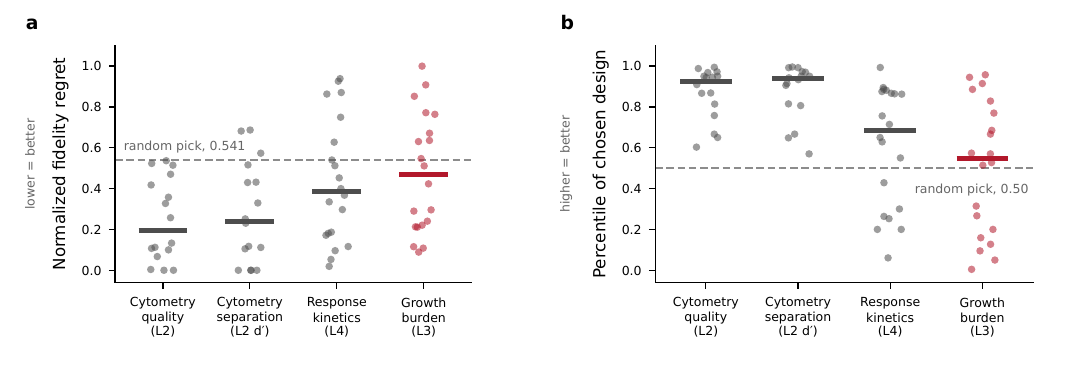}
  \caption{Exact fidelity regret with L1 as the reward. (a) Normalized regret for each
  (library, $k$, topology) cell; the dashed line is the L0 uniform-tie-break expectation,
  0.541. (b) Percentile of the L1-selected design under each reference. Panels c--d appear
  in Fig.~\ref{fig:fig1cd}.}
  \label{fig:fig1}
\end{figure}

Across the 72 enumerated single-axis cells with L1 as the reward, mean normalized regret is
0.369. Twenty-five cells (34.7\%) have regret at least 0.5, and 6 have zero regret. Median
normalized regret is 0.195 for cytometry quality, 0.241 for cytometry separation, and 0.467
for growth burden (Fig.~\ref{fig:fig1}a). The corresponding median percentiles of the L1
selection are 0.926, 0.938, and 0.549 (Fig.~\ref{fig:fig1}b); a uniform selection has median
percentile 0.5. The worst cell, \texttt{SC1C1G1T1}/\texttt{chain\_k3} against L3, has combination-space
normalized regret 0.9994 (0.45th percentile) and is also the most ordering-sensitive, falling
to 0.123 under exact ordered assignment; we treat it as an existence proof of severe regret
rather than a transferable magnitude (Appendix~\ref{app:ordered}).

\textbf{Regret depends on the validation axis.} Averaged over $k$,
\texttt{SC1C1G1T1} has mean normalized regret 0.113 against cytometry quality and 0.801
against growth burden. The pattern extends across libraries: all ten (library, $k$) burden
cells have positive regret. Under the prespecified operational rule in
Appendix~\ref{app:regret}, seven are UNSAFE, with worst regret above 0.35, and three are
CAUTION. The mean-field simulator does not represent metabolic load and selects gates that
the burden reference penalizes. Equation~\ref{eq:regret} therefore provides a pre-run screen
for the specific library and physical axis to be optimized.

\textbf{Topological verification ties the design space.} In all 120 supported L0 rows, the
number of designs tied at the optimum equals the number of valid designs. Uniform tie
breaking can therefore select any reference rank, giving a regret bracket $[0,1]$ and an
expected normalized regret of 0.541. These are properties of the tie distribution rather
than achieved optimization results.

The multiplicative verifier hierarchy of \citet{gencircuitrl} requires a valid document and
target truth table. Its topology-based functional check cannot rank gate assignments within
a fixed topology; its separate T8 evaluation instead introduces a Hill-response simulator.
An oracle whitelist gives an exact, non-implementable upper bound on the evaluated
per-gate-admissibility verifier class, subject to retaining at least 25\% of the design space.
Positive regret remains in 30 of 52 cells, including
17 of 20 burden cells, with a maximum of 0.511 (Appendix~\ref{app:verifier}). Thus this
verifier class cannot eliminate fidelity regret across the evaluated cells.

\section{Optimization pressure reaches the regret region}
\label{sec:opt}

\begin{figure}[t]
  \centering
  \includegraphics[width=0.52\textwidth]{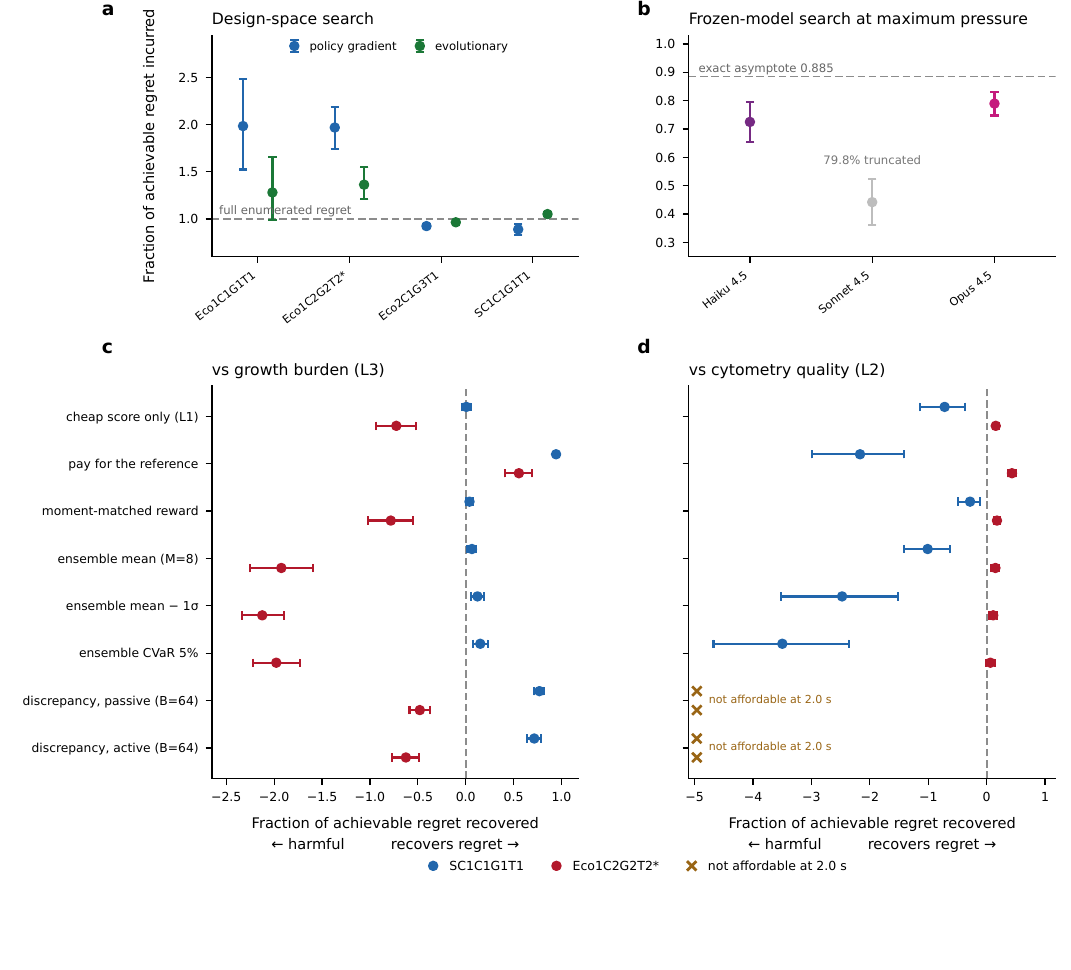}
  \caption{Optimization and mitigation results. Panels (a,b) show regret incurred; panels
  (c,d) show regret recovered on an axis unbounded below. (a) Design-space search, 30 seeds
  per arm. (b) Frozen-model search at maximum pressure; (c,d) Eight rewards at a matched modeled-cost 2.0\,s purse, 40 runs per
  cell. L2 discrepancy cells are marked \emph{not affordable}.}
  \label{fig:fig2}
\end{figure}

Exact regret characterizes the reward surface before optimization. We next test whether two
empirical optimization settings select the high-regret regions.

\textbf{Design-space search.} We apply REINFORCE with a baseline
\citep{williams1992} and $(\mu{+}\lambda)$ evolution \citep{beyer2002} to gate assignments
in 16 arms with 30 seeds per arm, evaluated against L3. The returned designs incur
0.789--3.252 of the achievable regret. All 16 arms have a 95\% CI lower bound of at least
0.70 (Appendix~\ref{app:stats}), and 7 have a lower bound of at least 1.0. Continued improvement of the cheap objective
can therefore return designs with lower reference scores than the enumerated cheap optimum.

\textbf{Frozen-model search.} Held-fixed public language models propose designs through the
same environment, without parameter updates (Appendix~\ref{app:repro}). Under methods that return the cheap score to
the model, regret rises from 0.602 [0.485, 0.718] at $R{=}1$ to 0.742 [0.647, 0.836] at
$R{=}5$ for iterative refinement, and from 0.605 at $G{=}1$ to 0.711 [0.613, 0.810] at
$G{=}4$ for evolutionary search. Pure best-of-$n$ shows no corresponding increase: Haiku
4.5 falls from 0.901 to 0.782, while Opus 4.5 changes from 0.880 to 0.885. The increase is
observed only when the cheap score is returned to the model.

\textbf{Model-tier comparison.} Paired over 44 cells at maximum pressure, the Opus 4.5 minus
Haiku 4.5 regret difference is $+0.053$ [$+0.001$, $+0.105$]. Opus also achieves a higher
cheap reward, with a difference of $+0.060$ [$+0.024$, $+0.096$], and reaches the exact
cheap optimum in 86.4\% of runs compared with 60.0\% for Haiku. On the two cells already
near the regret ceiling, the difference is $+0.012$ [$-0.032$, $+0.056$] ($n{=}30$), so
the aggregate effect is carried by cells with headroom. This direction matches Pan et al.'s
finding that capable agents attain a misspecified reward more effectively and incur greater
true-reward loss \citep{pan2022}. Sonnet 4.5 is a parsing artifact, not an intermediate
capability tier: 79.8\% of responses reach the output cap before completing a parseable
program.

\textbf{Matched falsification control.} On
\texttt{SC1C1G1T1}/\texttt{and\_k3}, identical proposals and cheap-reward selections are
scored against two references. Against L3, where enumerated regret is 0.771, observed regret
rises from 0.558 to 0.674 with pressure. Against L2\,$d'$, where enumerated regret is zero,
it falls from 0.313 to 0.199. The opposite trends under identical selection isolate
reference misspecification from a generic effect of optimization pressure (Appendix~\ref{app:pressure}).

\section{Mitigations at matched budget}
\label{sec:mit}
\label{sec:mit-neg}

\textbf{Discrepancy prediction transfers unevenly.} A discrepancy model predicts the
reference-minus-cheap score. For L2, gradient boosting attains $R^2$ values of 0.520--0.972
on random splits and $-1.334$--0.699 under shifts toward optimizer-preferred designs. Its
median $R^2$ on returned designs is 0.212, and on \texttt{SC1C1G1T1} its shifted RMSE is
34\% higher than using no correction. The L3 burden discrepancy remains more predictable,
with median $R^2=0.740$ on returned designs (Appendix~\ref{app:shift}).

The matched-budget study contains 1{,}280 runs at a budget of 2.0 modeled cost-equivalent
seconds, with 20 seeds and 2 optimizers per cell. The in-loop cost model sets the exchange
rates (Appendix~\ref{app:ladder}). Reference calls reduce the number of optimization steps
available within this budget. We report the fraction of achievable regret recovered (0 matches the cheap optimum, 1 reaches
the reference optimum, and negative values score below the cheap optimum; Appendix~\ref{app:mit}).

\textbf{Discrepancy-reward results.} Against L3 on \texttt{SC1C1G1T1}, a passively fitted
discrepancy reward ($B{=}64$) recovers 0.770 [0.716, 0.816], compared with 0.007 for the
unmitigated cheap reward ($p<10^{-4}$). On \texttt{Eco1C2G2T2}, its recovery remains
negative at $-0.479$ [$-0.587$, $-0.371$], although this is an improvement of $+0.245$ over
cheap-only ($p=0.023$). The active variant is indistinguishable from cheap-only in that cell
($p=0.41$), and active acquisition does not outperform passive acquisition at equal spend.
For L2, the 64-evaluation acquisition costs 4.05\,s and exceeds the 2.0\,s budget; the 160
affected runs are recorded as \texttt{budget\_infeasible} and have no recovery value.

\textbf{Reference-only results.} On \texttt{SC1C1G1T1} against L2, spending the 2.0\,s
budget on L2 recovers $-2.168$ [$-2.997$, $-1.418$], compared with $-0.720$ for the cheap
reward ($p<10^{-4}$). The budget buys about 32 evaluations, and neither optimizer leaves
initialization. On \texttt{Eco1C2G2T2}, reference-only recovery is 0.433 compared with
0.155 for cheap-only ($p<10^{-4}$). Aggregation would obscure this sign reversal, so the
results are reported by library.

\textbf{Risk-averse rewards.} Conservative objective models for offline model-based
optimization address the model misspecification targeted by these risk-averse arms
\citep{trabucco2021}. These rewards underperform the naive cheap reward in 3 of 4 cells. The
largest loss occurs for \texttt{SC1C1G1T1} against L2, where an 8-member
CVaR$_{0.05}$ reward recovers $-3.499$, compared with $-0.720$ for the naive reward.
Across 608 enumerated CVaR, mean--variance, and Wasserstein-DRO selections, 61 improve
regret, 137 worsen it, and 410 are unchanged. Ensemble spread has median absolute Spearman correlation 0.094 with reference shortfall (with
nominal L1 cost only 0.008 decades on \texttt{SC1C1G1T1}), and the positive training cell is
explained primarily by replacing the point estimate with an ensemble mean
(Appendix~\ref{app:robust}).

\textbf{Fidelity curricula.} On \texttt{SC1C1G1T1} against L3, each mixed schedule improves
on cheap-only recovery: the 50/50 and 0.25 switch schedules recover 0.783 and 0.890,
respectively, compared with 0.007. Reference-only reaches 0.944 and exceeds the mixed
schedules by $+0.053$ to $+0.775$. Against L2, the mixed schedules do not separate from
cheap-only. Reference-only recovery is $-2.256$; its paired difference against cheap-only is
$-1.536$ (Appendix~\ref{app:mit}).

\textbf{Fixed-budget re-scoring.} Rule-T allocates $\alpha{=}0.34$ of budget to re-score and
rerank top cheap designs; the fraction is selected and evaluated in-sample (no library held
out). Mean excess regret is 0.016 (cheap-only: 0.305; reference-only: 0.034), within 0.002 of
the leave-one-library-out oracle (0.014) and 0.006 of the leave-one-axis-out oracle (0.010).
It beats cheap-only in 330/360 cells and matches or improves on reference-only in 198/360
(161 wins, 37 ties; Appendix~\ref{app:decision}). Four adaptive methods track the best simple
policy within 0.106 at budgets of at least 128 reference calls; inferred fidelity-bias estimates
have max $|\rho_s|=0.293$ with enumerated regret (all $p>0.27$; Appendix~\ref{app:allocation}).

\section{Related work}

Prior work studies proxy failures via reward misspecification, hacking, and large-scale RLHF
reward-model overoptimization \citep{pan2022,amodei2016,skalse2022,gao2023}; our capability
ladder reproduces proxy gains with lower true reward. Conservative offline model-based objectives
motivate our risk-averse arms \citep{trabucco2021}; RLVR checkers reduce learned-model error but
leave physical performance simulator-ranked \citep{lambert2024}. Multi-fidelity optimization
and discrepancy correction, unreliable sources, and multi-fidelity best-arm identification
frame our pre-optimization regret measurement and tests of correction under optimizer-induced
distribution shift \citep{kennedy2001,forrester2007,kandasamy2016,kandasamy2017,mikkola2023,poiani2022}.
Cello/Cello 2.0 supply measured UCF libraries and expanded format; GenCircuit-RL supplies code
generation \citep{nielsen2016,jones2022,gencircuitrl}. Appendix~\ref{app:related} compares
multi-fidelity Bayesian optimization, bandits, distributionally robust optimization, and
Goodhart-style proxy optimization.

\section{Limitations}
\label{sec:limits}

Reference scores compose per-gate wet-lab measurements into circuit-level predictions. Without
whole-circuit wet-lab validation, composition error is unmeasured, and greater fidelity does not
establish ground truth. Appendix~\ref{app:regret} details range dependence and fit-residual-based
Hill-parameter uncertainty.

Five published Cello UCF libraries across three organisms are too few for statistical
attribution of regret to library properties (Appendix~\ref{app:shift}). The published
\texttt{Eco1C2G2T2} cytometry table has a decimal-point error, and \texttt{Bth1C1G1T1} lacks
single-cell cytometry; Appendix~\ref{app:ucf} reports correction sensitivity and unsupported-
axis handling.

Matched-budget results cover 2/4 cytometry-bearing libraries at a 2.0\,s modeled-cost-equivalent
budget. Enumeration is exact for combinations, not ordered assignments; the headline allocation
fraction is selected in-sample (Appendices~\ref{app:ordered} and~\ref{app:decision}). Our LM
experiments use frozen-weight inference-time search, not weight-space RL fine-tuning (e.g., GRPO
\citep{shao2024}).

\section{Impact statement}
\label{sec:impact}

Fidelity regret measures cheap simulators' reference shortfall before optimization and may reduce
selection of circuits with poor predicted reference performance. Standard published gate
characterizations support our intended-function study; verdicts apply only to evaluated models
and design space. SAFE means zero worst-case normalized regret there (Appendix~\ref{app:regret}),
not biological safety or experimentally validated reliability. Misplaced confidence is the risk:
syntax and topology may be verified while unmeasured simulator error misranks physically valid
designs.\looseness=-1

\bibliographystyle{plainnat}

\newpage
\section*{NeurIPS Paper Checklist}

\begin{enumerate}

\item {\bf Claims}
    \item[] Question: Do the main claims made in the abstract and introduction accurately reflect the paper's contributions and scope?
    \item[] Answer: \answerYes{}
    \item[] Justification: The abstract and introduction report the exact measurement over a 21,362-design combination subspace (\S\ref{sec:def}--\S\ref{sec:enum}, with ordered-assignment checks in Appendix~\ref{app:ordered}), the L2-versus-L3 asymmetry on \texttt{SC1C1G1T1} (0.113 versus 0.801), the L0 tie result and 25\%-floor oracle-whitelist bound, evidence from design-space and frozen-model optimization (\S\ref{sec:opt}), and matched-budget mitigations (\S\ref{sec:mit}). They report the fixed re-scoring rule as in-sample (mean excess regret 0.016 versus 0.305 for always-cheap and 0.034 for reference-only), and \S\ref{sec:limits} states that weight-space RL fine-tuning was not evaluated.

\item {\bf Limitations}
    \item[] Question: Does the paper discuss the limitations of the work performed by the authors?
    \item[] Answer: \answerYes{}
    \item[] Justification: \S\ref{sec:limits} is a dedicated limitations section. It states that the reference composes per-gate measurements without whole-circuit wet-lab validation; discusses range dependence and fit-residual Hill-parameter uncertainty (Appendix~\ref{app:regret}); notes the five-library scope (Appendix~\ref{app:shift}), the \texttt{Eco1C2G2T2} decimal defect, and \texttt{Bth1C1G1T1} missing cytometry (Appendix~\ref{app:ucf}); states the two-library 2.0\,s modeled-budget scope (Appendix~\ref{app:mit}); notes that enumeration is exact for the combination subspace rather than the ordered space and that the headline allocation fraction is selected in-sample (Appendices~\ref{app:ordered} and~\ref{app:decision}); and notes that weight-space RL fine-tuning was not evaluated.

\item {\bf Theory assumptions and proofs}
    \item[] Question: For each theoretical result, does the paper provide the full set of assumptions and a complete (and correct) proof?
    \item[] Answer: \answerNA{}
    \item[] Justification: The paper proves no theorems. The structural claim that L0 ties the design space is an empirical property of the enumerated cells (all 120 supported rows tie), reported as a measurement with its bracket $[0,1]$, not as a theorem.

    \item {\bf Experimental result reproducibility}
    \item[] Question: Does the paper fully disclose all the information needed to reproduce the main experimental results of the paper to the extent that it affects the main claims and/or conclusions of the paper (regardless of whether the code and data are provided or not)?
    \item[] Answer: \answerNo{}
    \item[] Justification: The paper specifies the ladder equations (Appendix~\ref{app:laddereq}), five public Cello UCF libraries with SHA-256 verification (Appendix~\ref{app:coverage}), the combination subspace, date-pinned public model ids (Appendix~\ref{app:repro}), optimizer settings, and statistical procedures (Appendix~\ref{app:stats}), but this submission omits the implementation, prompt templates, random seeds, and derived per-design tables needed to regenerate the reported results independently.

\item {\bf Open access to data and code}
    \item[] Question: Does the paper provide open access to the data and code, with sufficient instructions to faithfully reproduce the main experimental results, as described in supplemental material?
    \item[] Answer: \answerNo{}
    \item[] Justification: This submission includes the manuscript source and figures, while the analysis code, derived data tables, and frozen-model response logs are not distributed with the submission; the underlying Cello UCF libraries are publicly available (Appendix~\ref{app:coverage}).

\item {\bf Experimental setting/details}
    \item[] Question: Does the paper specify all the training and test details (e.g., data splits, hyperparameters, how they were chosen, type of optimizer) necessary to understand the results?
    \item[] Answer: \answerYes{}
    \item[] Justification: No model parameters were trained in this work; the frozen-model track changes no weights. Proposal counts, optimization-pressure levels for the three inference-time optimizers, cheap/reference level assignments per arm, matched-budget mitigation settings (Appendix~\ref{app:mit}), and the four discrepancy-model evaluation splits with their train/test construction (Appendix~\ref{app:shift}) are specified.

\item {\bf Experiment statistical significance}
    \item[] Question: Does the paper report error bars suitably and correctly defined or other appropriate information about the statistical significance of the experiments?
    \item[] Answer: \answerYes{}
    \item[] Justification: The stochastic optimizer results report 95\% percentile-bootstrap intervals over seeds or paired runs ($B{=}10{,}000$ and $B{=}2{,}000$, Appendix~\ref{app:stats}), with 30 seeds per design-space arm and 40 runs per matched-budget cell. The model-tier comparison is paired over 44 cells. Enumerated regret has no sampling interval because every design in each reported cell is evaluated.

\item {\bf Experiments compute resources}
    \item[] Question: For each experiment, does the paper provide sufficient information on the computer resources (type of compute workers, memory, time of execution) needed to reproduce the experiments?
    \item[] Answer: \answerYes{}
    \item[] Justification: Local enumeration and optimization ran CPU-only on an 8-core machine (7 parallel workers for enumeration at a $6.62\times$ aggregate speedup, Appendix~\ref{app:coverage}). Standalone and in-loop per-evaluation timings appear in Appendix~\ref{app:ladder}, and enumeration construction takes 11.9--21.7\,$\mu$s per design. The frozen-model proposals use date-pinned public API models with call counts, token counts (2.00M total tokens), and list-price spend (\$8.07) reported in Appendix~\ref{app:repro}.
    
\item {\bf Code of ethics}
    \item[] Question: Does the research conducted in the paper conform, in every respect, with the NeurIPS Code of Ethics \url{https://neurips.cc/public/EthicsGuidelines}?
    \item[] Answer: \answerYes{}
    \item[] Justification: The work involves no human subjects, personal data, or crowdsourcing. It uses published gate-characterization tables for laboratory chassis organisms and public API models. The paper reports the best-of-$n$ null result, the library-dependent discrepancy correction, budget-infeasible cells, and the high truncation rate of the intermediate model tier.

\item {\bf Broader impacts}
    \item[] Question: Does the paper discuss both potential positive societal impacts and negative societal impacts of the work performed?
    \item[] Answer: \answerYes{}
    \item[] Justification: \S\ref{sec:impact} is an impact statement. The positive case is that pre-optimization screening should reduce the selection of circuits whose predicted physical performance degrades under higher-fidelity models. Screening verdicts concern selection under the evaluated models, where SAFE means zero worst-case normalized regret in that space rather than biological safety. The stated risk is misplaced confidence: treating a topologically verified reward as validated when it retains unmeasured simulator error. The dual-use surface is discussed and assessed as limited because the work concerns whether intended circuits function using standard published gate characterizations.
    
\item {\bf Safeguards}
    \item[] Question: Does the paper describe safeguards that have been put in place for responsible release of data or models that have a high risk for misuse (e.g., pre-trained language models, image generators, or scraped datasets)?
    \item[] Answer: \answerNA{}
    \item[] Justification: No pretrained model, generative model, dataset, or software package is released with the submission.

\item {\bf Licenses for existing assets}
    \item[] Question: Are the creators or original owners of assets (e.g., code, data, models), used in the paper, properly credited and are the license and terms of use explicitly mentioned and properly respected?
    \item[] Answer: \answerNo{}
    \item[] Justification: The paper cites and credits the originating publications for all five public Cello UCF v2 gate libraries (Appendix~\ref{app:coverage}), the prior code-generation setting and topological verifier (\S1 and \S\ref{sec:enum}), and the date-pinned public API models (Appendix~\ref{app:repro}), but it does not explicitly state the applicable license terms or terms of service for those assets.

\item {\bf New assets}
    \item[] Question: Are new assets introduced in the paper well documented and is the documentation provided alongside the assets?
    \item[] Answer: \answerNA{}
    \item[] Justification: No new dataset, model, or software package is released with this submission.

\item {\bf Crowdsourcing and research with human subjects}
    \item[] Question: For crowdsourcing experiments and research with human subjects, does the paper include the full text of instructions given to participants and screenshots, if applicable, as well as details about compensation (if any)? 
    \item[] Answer: \answerNA{}
    \item[] Justification: No crowdsourcing and no research with human subjects.

\item {\bf Institutional review board (IRB) approvals or equivalent for research with human subjects}
    \item[] Question: Does the paper describe potential risks incurred by study participants, whether such risks were disclosed to the subjects, and whether Institutional Review Board (IRB) approvals (or an equivalent approval/review based on the requirements of your country or institution) were obtained?
    \item[] Answer: \answerNA{}
    \item[] Justification: No study participants.

\item {\bf Declaration of LLM usage}
    \item[] Question: Does the paper describe the usage of LLMs if it is an important, original, or non-standard component of the core methods in this research? Note that if the LLM is used only for writing, editing, or formatting purposes and does \emph{not} impact the core methodology, scientific rigor, or originality of the research, declaration is not required.
    \item[] Answer: \answerYes{}
    \item[] Justification: Large language models are a studied object of this work, not an incidental writing aid, and their role is reported in full. \S\ref{sec:opt} reports the frozen-model track: three date-pinned public models proposing designs through the environment with no weight access and no parameter updates. Appendix~\ref{app:repro} gives exact model ids, per-model call and token counts, list-price cost (\$8.07 total) and per-model proposal validity rates, including the 79.8\% truncation rate that makes one model's level a parsing artifact rather than a capability result.

\end{enumerate}

\newpage
\appendix
\section{Enumeration scope and per-library coverage}
\label{app:coverage}

The primary enumeration covers 21{,}362 designs across 5 measured Cello UCF libraries and 3
organisms. Each library was fetched from the CIDAR Cello-UCF repository and verified by
SHA-256. Table~\ref{tab:coverage} gives the available fidelity levels. Unsupported levels
are recorded as \emph{n.d.} with a reason and excluded from summary statistics.

The measurements originate from the initial \texttt{Eco1C1G1T1} Cello library
\citep{nielsen2016}, the \texttt{Eco1C2G2T2} plasmid library \citep{shin2020}, the
\texttt{Eco2C1G3T1} genome library \citep{park2020}, the yeast \texttt{SC1C1G1T1} library
\citep{chen2020}, and the \texttt{Bth1C1G1T1} \emph{B. thetaiotaomicron} library
\citep{taketani2020}. Cello 2.0 documents their common UCF representation
\citep{jones2022}.

\textbf{Combination subspace.} In the primary enumeration, a design is a $k$-element
combination of distinct gates evaluated once under the implementation's fixed slot-order
convention. The count is therefore
\[
2\sum_{n\in\{20,18,6,12,7\}}
\left[\binom{n}{3}+\binom{n}{4}\right]=21{,}362,
\]
where the factor of two is the two topologies. It is not the count of ordered injective
assignments, which would be
\[
2\sum_{n\in\{20,18,6,12,7\}}
\left[P(n,3)+P(n,4)\right]=432{,}372,
\qquad P(n,k)=\frac{n!}{(n-k)!}.
\]
Slot order can change circuit scores, particularly in cascades, so no permutation
invariance is assumed. The reported enumeration is exact for this combination subspace,
not for the full ordered-assignment space. At the measured 112.0\,ms aggregate CPU cost per
L2 evaluation, the ordered Phase-1 L2 column would require approximately 13.5 CPU-hours;
one 1{,}162{,}800-design Phase-2 library sweep would require approximately 1.5 CPU-days.
Appendix~\ref{app:ordered} reports exact and sampled ordered-assignment checks.

\begin{table}[h]
\centering
\small
\caption{Per-library coverage. \texttt{Bth1C1G1T1} ships growth-burden tables for all
seven gates but no single-cell cytometry, so the cytometry references (L2, L2\,$d'$ and
the joint L2+L3 references) are unsupported there; L3 and L4 are supported. The $\tau$
column counts gates with measured $\tau_{\text{on}}/\tau_{\text{off}}$ values;
$^{\dagger}$ denotes library-default kinetics.}
\label{tab:coverage}
\begin{tabular}{llrrrrrl}
\toprule
Library & Organism & Gates & Hill & Cytometry & Toxicity & $\tau$ & Levels \\
\midrule
\texttt{Eco1C1G1T1} & \emph{E.\ coli} NEB 10-beta & 20 & 20 & 20 & 20 & 0 & L0--L4$^{\dagger}$ \\
\texttt{Eco1C2G2T2} & \emph{E.\ coli} NEB 10-beta & 18 & 18 & 18 & 18 & 18 & L0--L4 \\
\texttt{Eco2C1G3T1} & \emph{E.\ coli} MG1655 K-12 & 6 & 6 & 6 & 6 & 0 & L0--L4$^{\dagger}$ \\
\texttt{SC1C1G1T1} & \emph{S.\ cerevisiae} BY4741 & 12 & 12 & 12 & 12 & 0 & L0--L4$^{\dagger}$ \\
\texttt{Bth1C1G1T1} & \emph{B.\ thetaiotaomicron} & 7 & 7 & \textbf{0} & 7 & 0 & L0,L1,L3,L4$^{\dagger}$ \\
\bottomrule
\end{tabular}
\end{table}

\textbf{L4 kinetics provenance.} Only \texttt{Eco1C2G2T2} supplies measured
$\tau_{\text{on}}/\tau_{\text{off}}$ values, for all 18 gates. The other four libraries
use default kinetics. Among the 20 enumerated L4 cells, the 4 measured-kinetics cells have
mean normalized regret 0.314 (maximum 0.452), while the 16 default-kinetics cells have mean
0.465 (maximum 0.938). L4 is therefore a common modeling axis but not a uniformly measured
one. L2 and L3 are composed from per-gate tables supplied by every library that supports
the corresponding level.

\textbf{Construction and scoring costs.} Constructing a design means enumerating an
assignment and building its circuit object without evaluating a ladder level. This costs
11.9--21.7\,$\mu$s per design; the largest single cell,
\texttt{Eco1C1G1T1}/\texttt{nor\_cascade\_k4} with ordered assignment, constructs 116{,}280
circuits in 2.33\,s. Reference scoring is measured separately. During parallel enumeration
with 7 workers, L2 uses 112.0\,ms of aggregate CPU time per design
(averaged over the 16 cytometry-bearing cells). Memory contention from propagating the 250-bin by 12-level
quantile table contributes to this parallel-enumeration cost.

The 112.0\,ms parallel-enumeration measurement, the 86.4\,ms standalone timing, and the
63.3\,ms optimization-loop cost model measure different execution contexts. Budget
comparisons use the optimization-loop model; construction timings are excluded. At the
enumerated problem sizes, 7-worker parallelism provides a $6.62\times$ aggregate speedup.

\section{Simulator ladder definitions}
\label{app:laddereq}

Let $\mathcal{P}=\{(0,0),(0,1),(1,0),(1,1)\}$ be the evaluated input patterns,
$t(p)\in\{0,1\}$ the target output, and
$\mathcal{P}^{+}=\{p:t(p)=1\}$ and $\mathcal{P}^{-}=\{p:t(p)=0\}$. Every ladder score is
maximized. No score uses an inverse-trigonometric function or requires a branch convention.

\textbf{L0: topology.} Boolean values are propagated through the regulatory graph, with a
NOR node emitting $\neg(\bigvee_i a_i)$. The dimensionless score is
\[
s_{\mathrm{L0}}=\frac{1}{|\mathcal{P}|}\sum_{p\in\mathcal{P}}
\mathbf{1}\!\left[b_{\mathrm{out}}(p)=t(p)\right]\in[0,1].
\]
It equals one when the topology implements the target and is independent of gate identity.

\textbf{L1: mean-field steady state.} Gate inputs are combined and propagated through the
fitted repressive Hill functions to obtain the output $y_{\mathrm{out}}(p)$ in relative
promoter units (RPU). Where specified by the library, the input combination includes the
fitted tandem-interference factor. The score is
\[
s_{\mathrm{L1}}=\log_{10}\!\left(
\frac{\min_{p\in\mathcal{P}^{+}}y_{\mathrm{out}}(p)}
     {\max_{p\in\mathcal{P}^{-}}y_{\mathrm{out}}(p)}\right),
\]
measured in decades of RPU separation. Positive values place every ON output above every
OFF output.

\textbf{L2: cytometry quality and separation.} A gate's single-cell measurements provide
$P(y\mid x_\ell)$ at 12 input levels. For an input distribution $Q$, propagation uses the
equal-weight quantile approximation
\[
P(y)=\int P(y\mid x)Q(x)\,\mathrm{d}x
\approx \frac{1}{n_{\mathrm{quad}}}\sum_{j=1}^{n_{\mathrm{quad}}}P(y\mid x_j),
\qquad x_j=Q^{-1}(u_j).
\]
This composition assumes that the measured conditional spread is intrinsic to each gate
and independent between gates in a cascade. The ON and OFF pattern distributions are pooled
separately. If $\mu$ and $\sigma$ denote their means and standard deviations in
$\log_{10}$ RPU, then
\[
s_{\mathrm{L2}}=(\mu_{\mathrm{on}}-\mu_{\mathrm{off}})
-(\sigma_{\mathrm{on}}+\sigma_{\mathrm{off}}),
\qquad
s_{\mathrm{L2}\,d'}=
\frac{|\mu_{\mathrm{on}}-\mu_{\mathrm{off}}|}
{\sqrt{(\sigma_{\mathrm{on}}^2+\sigma_{\mathrm{off}}^2)/2}}.
\]
The first score is in decades of RPU; $d'$ is dimensionless. The absolute value in the
$d'$ numerator makes it invariant to reversing the ON/OFF mean ordering, so it certifies
separation but not analog polarity. The signed L1 and L2 scores retain polarity and become
negative when that ordering inverts. We use $d'$ only as a separation reference, never as the
sole certificate of correct direction.

\textbf{L3: growth burden.} The toxicity table gives each gate's relative growth rate
$g_i(x)$ at input level $x$. Independent relative-growth effects are multiplied, and the
worst truth-table state is scored:
\[
s_{\mathrm{L3}}=\min_{p\in\mathcal{P}}
\prod_{i\in\mathrm{gates}}g_i\!\left(x_i(p)\right)\in(0,1],
\qquad
\mathrm{penalty}_{\log_{10}}=-\log_{10}s_{\mathrm{L3}}.
\]
Each $x_i(p)$ is the L1 mean-field input to gate $i$.

\textbf{L4: response kinetics.} Each logic node relaxes toward its Hill steady state as
\[
\frac{\mathrm{d}y_i}{\mathrm{d}t}
=\frac{\operatorname{Hill}_i(x_i(t))-y_i}{\tau_i}.
\]
The implementation integrates this system with LSODA
\citep{hindmarsh1983,petzold1983} and measures the output's 10--90\% response time
$t_{\mathrm{resp}}$ for the configured OFF-to-ON input step. The maximize score is
\[
s_{\mathrm{L4}}=-\log_{10}\!\left(\max(t_{\mathrm{resp}},10^{-6})\right),
\]
the negative logarithm of response time in simulator time units. Direction-dependent
$\tau_i$ values come from the UCF where available and from a library default otherwise.

\textbf{Joint L2+L3 references.} The three joint references use
\[
s_{\mathrm{joint}}=s_{\mathrm{L2}}-w\,\mathrm{penalty}_{\log_{10}},
\qquad w\in\{0.5,1,2\}.
\]
Both terms are measured in decades, so $w$ is dimensionless.

\textbf{Missing data.} An unsupported level is neither imputed nor scored as zero.
\texttt{Bth1C1G1T1} lacks cytometry, so L2, L2\,$d'$, and the three joint references are
recorded as \emph{n.d.} and excluded from summaries. Default kinetics at L4 are retained but
identified separately in Table~\ref{tab:coverage}.

\section{Full combination-subspace exact-regret tables}
\label{app:regret}

\begin{figure}[h]
  \centering
  \includegraphics[width=\textwidth]{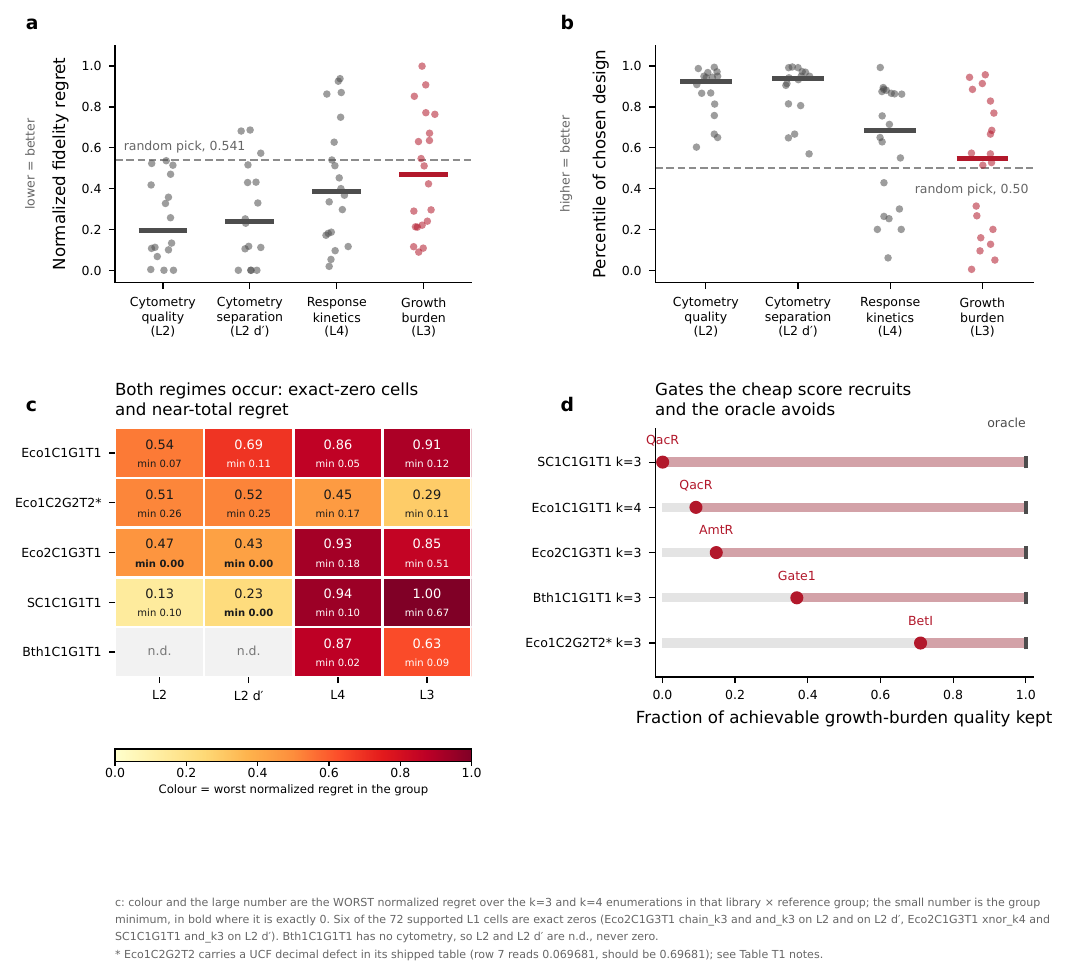}
  \caption{Full four-panel version of Fig.~\ref{fig:fig1}. (c) Worst normalized regret per
  library~$\times$~reference with the group minimum beneath, bold where it is exactly zero;
  \texttt{Bth1C1G1T1} cytometry cells are \emph{n.d.} because cytometry is unsupported.
  (d) Fraction of achievable
  growth-burden quality kept on each library's worst cell, annotated with the gate the cheap
  score recruits and the oracle avoids.}
  \label{fig:fig1cd}
\end{figure}

The full combination-subspace enumeration spans 280 (library, topology, cheap simulator,
reference) rows, recording the selected cheap optimum, oracle design, number of cheap-score
ties, reference distribution, and rank agreement. The 72 main-text cells are its evaluable
single-axis L1-reward subset. Table~\ref{tab:worstcell} reports the worst normalized regret
per (library, reference) over $k \in \{3,4\}$, as visualized in Fig.~\ref{fig:fig1}c.

\textbf{Regret metrics.} Three quantities use different denominators. Let
$s^{\max}=\max_d s_{\mathrm{ref}}(d)$, $s^{\min}=\min_d s_{\mathrm{ref}}(d)$, and let
$d^*_{\mathrm{cheap}}$ be the cheap-score maximizer. For a returned design $d$, define
\[
R_{\mathrm{range}}(d)=
\frac{s^{\max}-s_{\mathrm{ref}}(d)}{s^{\max}-s^{\min}},
\qquad
R_{\mathrm{frac}}(d)=
\frac{s^{\max}-s_{\mathrm{ref}}(d)}
     {s^{\max}-s_{\mathrm{ref}}(d^*_{\mathrm{cheap}})},
\qquad
\mathrm{Rec}(d)=1-R_{\mathrm{frac}}(d).
\]
$R_{\mathrm{range}}$ is the range-normalized enumeration statistic and lies in $[0,1]$.
$R_{\mathrm{frac}}$ is the regret fraction incurred relative to the cheap optimum: it is
one at $d^*_{\mathrm{cheap}}$ and can exceed one for a design that the reference scores
worse. $\mathrm{Rec}$ is the recovered fraction in Appendix~\ref{app:mit}; it is zero at
the cheap optimum, one at the reference optimum, and unbounded below. The enumeration fixes
both anchors. When the cheap optimum is also reference-optimal, as in 6 of the 72 main-text
cells, the denominator of $R_{\mathrm{frac}}$ is zero; $R_{\mathrm{frac}}$ and
$\mathrm{Rec}$ are recorded as \texttt{nan} and excluded from means with the excluded count
reported. The matched zero-regret control in Appendix~\ref{app:pressure} instead reports
$R_{\mathrm{range}}$ for optimizer-returned designs, which can be positive when finite
optimization does not return the exact cheap optimum.

\textbf{Cheap-score ties.} If the cheap argmax contains several designs with different
reference scores, Eq.~\ref{eq:regret} requires an explicit tie rule. Across the enumeration,
all 120 L1 cells have a unique cheap argmax, so tie-breaking does not affect any reported L1
point estimate. All 120 L0 cells are fully tied, with 15--4{,}845 optima, and are reported
with the interval from best-tie to worst-tie regret and the uniform-tie-break expectation.

We also evaluate sensitivity to near-ties within a tolerance of $10^{-3}$ decades of L1 score.
Forty-nine of 120 L1 cells contain a
near-tied competitor; in 41, choosing among near-tied designs changes absolute regret by
more than 0.01. The median normalized spread is 0.126 and the maximum is 0.697 for
\texttt{Eco1C1G1T1}/\texttt{and\_k3} against L3.

\begin{table}[h]
\centering
\small
\caption{Worst normalized regret per library and reference axis, L1 paying the reward.
\emph{n.d.} = the reference is not evaluable for that library.
$^{*}$ carries the \texttt{Eco1C2G2T2} UCF defect flag (\S\ref{app:ucf}).}
\label{tab:worstcell}
\begin{tabular}{lrrrr}
\toprule
Library & L2 (cyt.\ quality) & L2\,$d'$ (cyt.\ sep.) & L3 (burden) & L4 (kinetics) \\
\midrule
\texttt{Eco1C1G1T1} & 0.536 & 0.687 & 0.908 & 0.863 \\
\texttt{Eco1C2G2T2}$^{*}$ & 0.514 & 0.516 & 0.290 & 0.452 \\
\texttt{Eco2C1G3T1} & 0.470 & 0.431 & 0.852 & 0.926 \\
\texttt{SC1C1G1T1} & 0.133 & 0.230 & \textbf{0.999} & 0.938 \\
\texttt{Bth1C1G1T1} & \emph{n.d.} & \emph{n.d.} & 0.630 & 0.870 \\
\bottomrule
\end{tabular}
\end{table}

\textbf{Reference provenance.} The UCF provides cytometry, Hill fits, and toxicity for
individual gates; the ladder composes these measurements into circuit-level predictions.
The study contains no whole-circuit wet-lab validation, and composition error is therefore
unmeasured. A higher rung is a more detailed model rather than ground truth.

Hill-parameter uncertainty is derived from cytometry fit residuals and represents uncertainty
in the fitted parameters. The UCF provides one median per input level, so replicate-level
measurement scatter cannot be recovered. The optimizer study uses the AND target at
$k{=}4$, whereas the enumeration includes other targets. Results are kept separate across
targets and, in the mitigation study, across libraries because the observed effects can
reverse sign (Appendix~\ref{app:mit}).

\textbf{Range dependence.} Normalized regret divides by the reference range over the
enumerated space, so a narrow range inflates it. \texttt{Bth1C1G1T1}/\texttt{and\_k3}
against L4 is 0.870 normalized but only 0.085 decades in absolute terms. Each normalized
figure in this paper is paired with either the percentile of the pick or
the absolute value.

\textbf{Operational verdict rule.} The prespecified screening convention labels a (library, $k$, reference) cell SAFE when its
worst normalized regret is 0, CAUTION when it is in $(0,0.35]$, UNSAFE when it exceeds
0.35, and \emph{n.d.} when the reference is unsupported. The 0.35 boundary is an
operational triage threshold and carries no significance-test interpretation. Among 70 rows, 41 are UNSAFE, 14
CAUTION, 5 SAFE, and 10 \emph{n.d.}; the L3 burden axis contains 7 UNSAFE and 3 CAUTION
rows.

\section{Level disagreement and the cost ladder}
\label{app:ladder}

\begin{figure}[h]
  \centering
  \includegraphics[width=\textwidth]{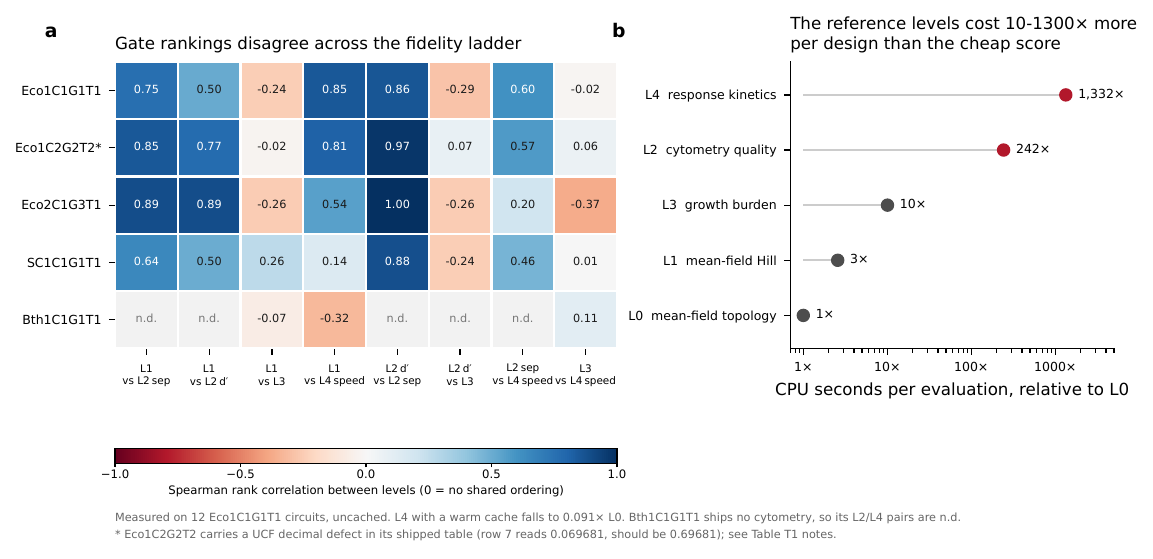}
  \caption{(a) Spearman rank correlation between fidelity levels on each library's gate
  set. (b) Cost per evaluation relative to L0, log axis.}
  \label{fig:figA1}
\end{figure}

The fidelity levels induce different gate orderings. L1-versus-L3 Spearman correlations are
$-0.243$ for \texttt{Eco1C1G1T1}, $-0.024$ for \texttt{Eco1C2G2T2}, $-0.257$ for
\texttt{Eco2C1G3T1}, $-0.071$ for \texttt{Bth1C1G1T1}, and $+0.260$ for
\texttt{SC1C1G1T1}. Four are negative, and all are close to zero. Thus L1 provides little
information about the measured burden ordering, which accounts for the burden regret in
\S\ref{sec:enum}. L1-versus-cytometry agreement is higher, ranging from 0.50 to 0.89.

\textbf{Standalone and optimization-loop costs.} The standalone benchmark times isolated
evaluations across L0--L4. The optimization-loop model measures L0--L3 calls in the training
execution context and determines every budget comparison.

\begin{table}[h]
\centering
\small
\caption{The two measured cost bases. \textbf{Every budget comparison in this paper uses
the in-loop exchange rates}; the standalone benchmark supplies the relative-to-L0 ladder in Fig.~\ref{fig:figA1}b.}
\label{tab:cost}
\begin{tabular}{llrrr}
\toprule
Timing basis & Measures & L2 (s) & L2/L1 & L3/L1 \\
\midrule
Standalone benchmark & isolated per-evaluation, all 5 levels & 0.086403 & 94.3 & 3.9 \\
In-loop cost model & \textbf{optimization-loop execution, L0--L3} & 0.063297 & \textbf{67.4} & \textbf{2.2} \\
\bottomrule
\end{tabular}
\end{table}

The standalone benchmark on 12 \texttt{Eco1C1G1T1} circuits gives L0 0.000357\,s, L1
0.000916\,s, L3 0.003585\,s, L2 0.086403\,s, L4 0.475384\,s, and cached L4 0.000033\,s. It
is the only measurement covering L4. The optimization-loop model gives L0 0.000418\,s, L1
0.000939\,s, L2 0.063297\,s, and L3 0.002053\,s. The measurements agree within 2.5\% on L1
and 15\% on L0; execution-context overhead and amortization produce differences of
1.37$\times$ on L2 and 1.75$\times$ on L3. All optimization budgets use the corresponding
exchange rates of 67.4 L1 calls per L2 call and 2.2 L1 calls per L3 call.

\textbf{Ensemble cost.} Under the optimization-loop model, one CPU-hour supports 59{,}903
evaluations of a 64-member L1 ensemble or 56{,}874 direct L2 evaluations. The break-even
ensemble size is 67 members, equal to the L2/L1 exchange rate. An 8-member ensemble provides
479{,}228 evaluations per CPU-hour. The standalone measurement would place break-even at
94 members, demonstrating why budget comparisons use a single execution context. In
\S\ref{sec:mit}, all variants select from identical proposal sets and match the
high-fidelity budget by construction.

\section{Discrepancy models under distribution shift}
\label{app:shift}

\begin{figure}[h]
  \centering
  \includegraphics[width=\textwidth]{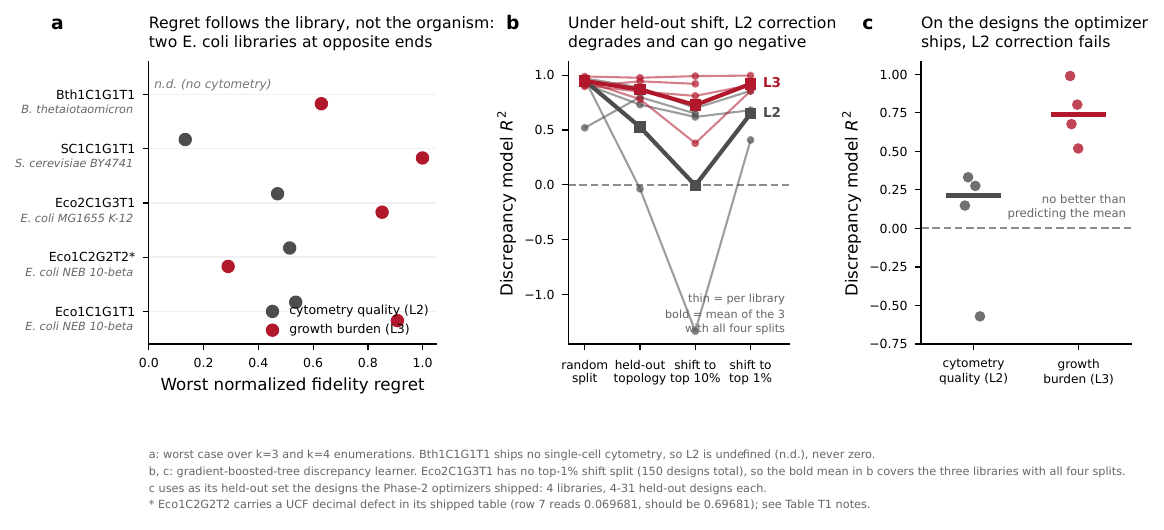}
  \caption{(a) Worst normalized regret per library on cytometry and burden; two
  \emph{E.\ coli} libraries land at opposite ends. (b) Discrepancy-model $R^2$ across four
  splits: L2 includes negative values under shift, whereas L3 remains positive. (c) $R^2$
  on optimizer-returned designs: L2 median 0.212, L3 median 0.740.}
  \label{fig:fig3}
\end{figure}

In Fig.~\ref{fig:fig3}a, the two \emph{E.\ coli} NEB 10-beta libraries have worst burden
regrets of 0.908 and 0.290, while libraries from three organisms interleave in the observed
ordering. Across these five libraries, regret varies more between gate libraries than by host
organism, although five libraries are too few for formal cross-library attribution.

Four splits were evaluated per (library, reference, learner): \texttt{random},
\texttt{topology\_holdout}, \texttt{shift\_q90} (train below the 90th cheap-score
percentile, test above), and \texttt{shift\_q99}. A fifth evaluation scores the model on
the optimizer-returned designs from the design-space search.

\begin{table}[h]
\centering
\small
\caption{Gradient-boosted discrepancy-model $R^2$, range across libraries. L2 includes
negative values under distribution shift; L3 remains positive.}
\label{tab:shift}
\begin{tabular}{lll}
\toprule
Split & L2 cytometry quality & L3 growth burden \\
\midrule
\texttt{random} & 0.520 -- 0.972 & 0.898 -- 0.988 \\
\texttt{topology\_holdout} & $-0.036$ -- 0.887 & 0.780 -- 0.976 \\
\texttt{shift\_q90} & $\mathbf{-1.334}$ -- 0.699 & 0.379 -- 0.991 \\
\texttt{shift\_q99} & 0.409 -- 0.857 & 0.856 -- 0.996 \\
\midrule
optimizer-returned (median) & \textbf{0.212} & \textbf{0.740} \\
\bottomrule
\end{tabular}
\end{table}

On \texttt{SC1C1G1T1} under \texttt{shift\_q90}, the L2 correction's RMSE reduction relative
to no correction is $-0.338$, corresponding to a 34\% increase in RMSE. On the designs
returned by the optimizers, it reaches $R^2=-0.571$ for the same library. The L3 correction
has positive $R^2$ for every library and split. Feature-importance analysis
attributes the L3 result to two directly
available features, reducing its dependence on extrapolated distributional properties.

\textbf{Sample sizes.} \texttt{Eco2C1G3T1} contains 150 designs and has no
\texttt{shift\_q99} split. The three remaining cytometry-bearing libraries carry all four
splits, and \texttt{Eco2C1G3T1} is excluded from means that require all four. Test sets of
optimizer-returned designs contain 4--31 unique designs, so these estimates have substantial
sampling uncertainty.

\section{Pressure--regret trajectories}
\label{app:pressure}

\begin{figure}[h]
  \centering
  \includegraphics[width=\textwidth]{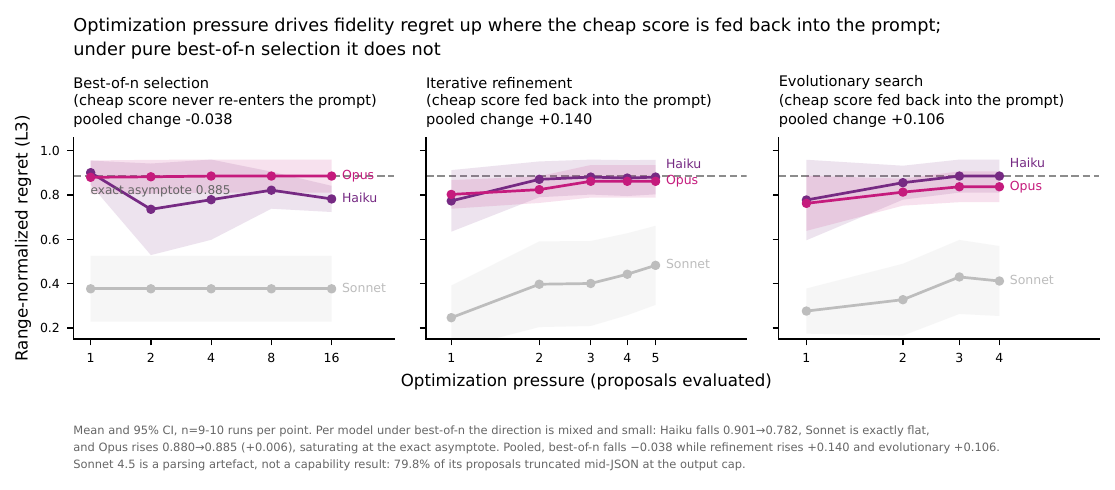}
  \caption{Range-normalized regret against optimization pressure for the frozen models. Regret
  rises under optimizers that feed the cheap score back into the prompt and does
  \emph{not} rise under pure best-of-$n$ selection. Range-normalized regret $R_{\mathrm{range}}\in[0,1]$
  (\S\ref{app:regret}) remains well defined on the zero-enumerated-regret cell of the falsification
  control, where a cheap-optimum-relative fraction has a zero denominator.}
  \label{fig:figA5}
\end{figure}

The values in this section are $R_{\mathrm{range}}$, not $R_{\mathrm{frac}}$. Under
best-of-$n$, Haiku 4.5 regret falls from 0.901 to 0.782, Opus 4.5 changes from 0.880
to 0.885, and Sonnet 4.5 is constant. Pooled across models, iterative-refinement regret is
0.602, 0.691, 0.714, 0.727, and 0.742 for $R=1\ldots5$; evolutionary-search regret is
0.605, 0.665, 0.718, and 0.711 for $G=1\ldots4$. Regret increases when the optimizer feeds
the cheap score back to the model and shows no corresponding increase under best-of-$n$.

\textbf{Matched falsification control.} On the matched cell
\texttt{SC1C1G1T1}/\texttt{and\_k3}, identical proposals under identical cheap-reward
selection scored against two references: against L3 (enumerated regret 0.771) regret
rises 0.558, 0.575, 0.623, 0.628, 0.674 with pressure; against L2\,$d'$ (enumerated regret
exactly 0.000) it falls 0.313, 0.291, 0.248, 0.233, 0.199. The opposite trends for the same
proposals and selection rule identify reference misspecification as the source of the effect.

\section{Library data and coverage sensitivity}
\label{app:ucf}

\textbf{Eco1C2G2T2 decimal point.} Row 7 of the published cytometry table reads
$0.069681$; $0.69681$ restores monotonicity of the transfer curve. Both variants were
enumerated over all 30{,}600 designs. The cheap argmax is unchanged (index 6350), as is
the number of cheap ties (2). This 30{,}600-design sensitivity enumeration is
reported separately from the 21{,}362-design primary enumeration summarized in the main
text.

\begin{table}[h]
\centering
\small
\caption{\texttt{Eco1C2G2T2} sensitivity to the decimal defect. The L3 result is
unaffected; the L2 result changes materially, so both are reported.}
\label{tab:ucf}
\begin{tabular}{llrrr}
\toprule
Variant & Reference & Absolute regret & Percentile of pick & Spearman \\
\midrule
as published & L2 & 1.176 & 64.8 & 0.531 \\
corrected & L2 & 0.879 & 84.3 & 0.593 \\
as published & L3 & 0.166 & 97.1 & 0.063 \\
corrected & L3 & 0.166 & 97.1 & 0.063 \\
\bottomrule
\end{tabular}
\end{table}

\textbf{Bth1C1G1T1 cytometry coverage.} This library provides no single-cell cytometry
collection. L2, L2\,$d'$, and the three joint L2+L3 references are unsupported and appear
as \emph{n.d.} in figures and tables. L3 and L4 remain supported for the enumeration; its
worst L4 cell is the 0.870-normalized, 0.085-decade example above. A value of zero would be
semantically incorrect: the L1/L2 score $\log_{10}(\min_{\text{ON}}/\max_{\text{OFF}})$
equals zero when ON and OFF are equal, which is higher than most of this design space.

\section{Matched-budget mitigation study}
\label{app:mit}

\textbf{Design.} The study contains 1{,}280 runs: 2 libraries, 2 reference axes, 8 reward
variants, and 40 runs per cell (20 seeds for each of REINFORCE and
$(\mu{+}\lambda)$ evolution). Every run receives a modeled budget equivalent to 2.0\,s under
the optimization-loop cost model, so reference calls reduce the available training steps. The
reported quantity is the fraction of achievable regret recovered, $1-$ regret fraction.
Zero corresponds to the cheap simulator's selected optimum, one to the enumerated reference
optimum, and negative values to designs below the cheap optimum under the reference. The
scale is unbounded below.

\begin{table}[h]
\centering
\small
\caption{Fraction of achievable regret recovered at a matched modeled budget equivalent to
2.0\,s, mean [95\% CI]
over 40 runs. \emph{infeasible} indicates that the declared acquisition cost exceeds the
full budget and permits no training calls.}
\label{tab:mit}
\begin{tabular}{lrrrr}
\toprule
& \multicolumn{2}{c}{\texttt{SC1C1G1T1}} & \multicolumn{2}{c}{\texttt{Eco1C2G2T2}} \\
\cmidrule(lr){2-3}\cmidrule(lr){4-5}
Reward variant & L3 burden & L2 cytometry & L3 burden & L2 cytometry \\
\midrule
cheap L1 (naive)          & 0.007 & $-0.720$ & $-0.724$ & 0.155 \\
reference simulator       & \textbf{0.944} & $\mathbf{-2.168}$ & 0.555 & 0.433 \\
ensemble mean, $M{=}8$    & 0.065 & $-1.011$ & $-1.925$ & 0.149 \\
ensemble mean $-1\sigma$  & 0.123 & $-2.475$ & $-2.124$ & 0.111 \\
ensemble CVaR$_{0.05}$    & 0.152 & $\mathbf{-3.499}$ & $-1.978$ & 0.063 \\
moment matched            & 0.040 & $-0.285$ & $-0.782$ & 0.176 \\
discrepancy, passive $B{=}64$ & \textbf{0.770} & \emph{infeasible} & $-0.479$ & \emph{infeasible} \\
discrepancy, active $B{=}64$  & 0.716 & \emph{infeasible} & $-0.624$ & \emph{infeasible} \\
\bottomrule
\end{tabular}
\end{table}

\textbf{Budget infeasibility.} The discrepancy variants allocate 64 high-fidelity
evaluations from the same budget. For L2, this acquisition costs 4.05\,s and exceeds the
2.0\,s budget. The 160 affected runs are recorded as \texttt{budget\_infeasible} with NaN
recovery because they execute no training calls. Smaller L2 acquisition budgets were not
evaluated.

\textbf{Discrepancy reward.} On \texttt{SC1C1G1T1} against L3, passive discrepancy
correction recovers 0.770, an improvement of $+0.763$ over cheap-only ($p<10^{-4}$). On
\texttt{Eco1C2G2T2} against L3, its recovery is $-0.479$ [$-0.587$, $-0.371$], an
improvement of $+0.245$ over cheap-only ($p=0.023$). The active variant differs from
cheap-only by $+0.100$ ($p=0.41$). These library-specific results are reported separately.

\textbf{Reference-only reward.} On
\texttt{SC1C1G1T1} against L2, the L2 reference-only arm recovers $-2.168$
[$-2.997$, $-1.418$], worse than the naive cheap reward's $-0.720$ ($\Delta = -1.448$,
$p<10^{-4}$ paired), because 2\,s buys roughly 32 L2 evaluations, not enough for either
optimizer to leave its initialization. On \texttt{Eco1C2G2T2}, recovery is 0.433
[0.361, 0.502] against cheap-only's 0.155
($\Delta = +0.277$, $p<10^{-4}$). Against L3, where evaluations are far cheaper, it recovers
0.944 on \texttt{SC1C1G1T1} and 0.555 on \texttt{Eco1C2G2T2}. The L2 sign reversal shows
that the value of reference-only optimization at fixed budget depends on the library.

\textbf{Risk aversion.} Cheap-ensemble risk measures lose to the naive cheap reward in 3 of
the 4 cells; the exception is \texttt{SC1C1G1T1}\,$\times$\,L3, where CVaR$_{0.05}$ gives a
small gain (0.152 versus 0.007). The nominal-performance premium these rewards pay is
small: 0.008 decades of L1 score on \texttt{SC1C1G1T1} across $\kappa=0.5$--2 and 0.340 on
\texttt{Eco1C2G2T2}. Here $\kappa$ is the risk-aversion multiplier in the
mean$-{}\kappa\sigma$ statistic and is unrelated to circuit size
$k\in\{3,4\}$. The losses arise because the predictive spread has little alignment
with reference shortfall (Appendix~\ref{app:robust}).

\begin{table}[h]
\centering
\small
\caption{Curriculum schedules on \texttt{SC1C1G1T1} at the matched budget. The
\texttt{reference\_only} row provides the direct-reference comparator.}
\label{tab:curr}
\begin{tabular}{lrrr}
\toprule
Schedule & switch & L3 recovered [95\% CI] & vs \texttt{cheap\_only} ($p$) \\
\midrule
\texttt{cheap\_only}           & --   & 0.007 [$-0.036$, 0.050] & -- \\
\texttt{reference\_then\_cheap} & 0.50 & 0.169 [0.093, 0.250] & $+0.162$ ($<10^{-4}$) \\
\texttt{interleaved}           & 0.50 & 0.641 [0.540, 0.739] & $+0.634$ ($<10^{-4}$) \\
\texttt{cheap\_then\_reference} & 0.75 & 0.618 [0.506, 0.733] & $+0.611$ ($<10^{-4}$) \\
\texttt{cheap\_then\_reference} & 0.50 & 0.783 [0.697, 0.861] & $+0.776$ ($<10^{-4}$) \\
\texttt{cheap\_then\_reference} & 0.25 & \textbf{0.890} [0.853, 0.926] & $+0.883$ ($<10^{-4}$) \\
\midrule
\texttt{reference\_only} (ceiling) & -- & \textbf{0.944} [0.926, 0.961] & $+0.937$ ($<10^{-4}$) \\
\midrule
\multicolumn{4}{l}{\emph{Same schedules against L2:} no mixed schedule separates from
cheap-only (best $+0.261$, $p{=}0.13$);} \\
\multicolumn{4}{l}{\texttt{reference\_only} recovery is $-2.256$; its paired difference} \\
\multicolumn{4}{l}{from \texttt{cheap\_only} is $-1.536$ ($p{=}0.0002$).} \\
\bottomrule
\end{tabular}
\end{table}

\begin{figure}[h]
  \centering
  \includegraphics[width=\textwidth]{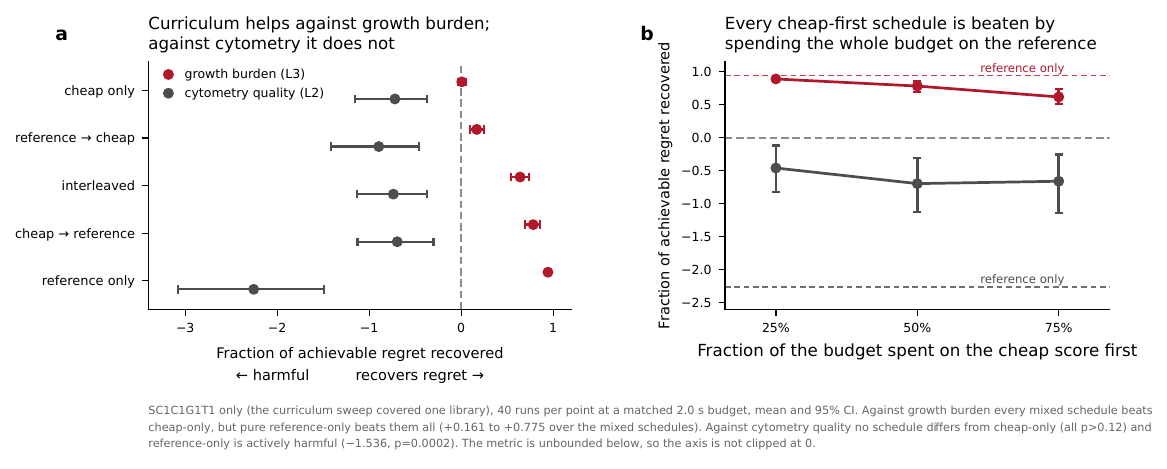}
  \caption{Fidelity curricula on \texttt{SC1C1G1T1} (the only library swept), 40 runs per
  point at the matched modeled budget equivalent to 2.0\,s. (a) Against growth burden every mixed schedule beats
  cheap-only; against cytometry quality none does, and \texttt{reference\_only} has negative
  recovery. (b) On growth burden (L3), every cheap-first schedule lies below the
  \texttt{reference\_only} line. On cytometry quality (L2), the ordering inverts because
  \texttt{reference\_only} is harmful there.
  The recovery axis is unbounded below.}
  \label{fig:figA6}
\end{figure}

\textbf{Curriculum.} The sweep covers \texttt{SC1C1G1T1}. Against L3,
cheap-then-reference at a 50/50 split recovers 0.783 compared with 0.007 for cheap-only;
the reference-then-cheap control recovers 0.169, and a 0.25 switch recovers 0.890.
Reference-only recovers 0.944 and exceeds the mixed schedules by $+0.053$ to $+0.775$.
Against L2, no mixed schedule separates from cheap-only: the largest gain is $+0.261$
[$-0.071$, 0.621] at a 0.25 switch ($p=0.13$), and the matched 50/50 difference is
$+0.025$ ($p=0.88$). Reference-only recovery is $-2.256$ [$-3.075$, $-1.493$]; its paired
difference from cheap-only is $-1.536$ [$-2.327$, $-0.789$] ($p=0.0002$).

\textbf{Reference-only comparators.} The $-1.536$ value is a paired difference within the
curriculum sweep, whereas the $-2.168$ in Table~\ref{tab:mit} is an absolute recovery level
from the mitigation study. The comparable curriculum-sweep level is $-2.256$. The remaining
difference of approximately 0.09 between the two levels reflects budget accounting rather
than different seeds: both use the same 20 seeds and two optimizers, and 32 of 40 paired
runs are identical. The curriculum harness stops after 31 reference evaluations at
1.962\,s charged cost, while the mitigation harness admits a 32nd evaluation and records
2.026\,s. The values are therefore reported separately.

\textbf{Scope.} The study covers 2 of the 4 cytometry-bearing libraries
(\texttt{SC1C1G1T1} and \texttt{Eco1C2G2T2}) at a matched modeled budget equivalent to
2.0\,s. Absolute recovery
values, especially for the L2 reference-only arm, are specific to this budget.

\textbf{Controls.} A reference-blinding check is run for every variant that does not read the
reference: multiplying reference scores by $-100$ leaves the optimization trajectory unchanged.
The discrepancy reward reads reference scores only on the designs in its paid acquisition set,
and perturbing reference scores outside that set leaves the returned design unchanged.

\section{Models, tokens and reproduction cost}
\label{app:repro}

The frozen-model experiment used three date-pinned public model ids:
\texttt{claude-haiku-4-5-20251001}, \texttt{claude-sonnet-4-5-20250929} and
\texttt{claude-opus-4-5-20251101}.

\begin{table}[h]
\centering
\small
\caption{Frozen-model spend at public list prices. Totals are 511 fresh calls,
2.00M tokens, \$8.07.}
\label{tab:spend}
\begin{tabular}{lrrrr}
\toprule
Model & Fresh calls & Input tokens & Output tokens & List cost (USD) \\
\midrule
Haiku 4.5 & 182 & 840{,}626 & 95{,}844 & 1.32 \\
Opus 4.5 & 164 & 760{,}515 & 56{,}453 & 5.21 \\
Sonnet 4.5 & 165 & 182{,}683 & 66{,}000 & 1.54 \\
\midrule
\textbf{Total} & \textbf{511} & \textbf{1{,}783{,}824} & \textbf{218{,}297} & \textbf{8.07} \\
\bottomrule
\end{tabular}
\end{table}

The frozen-model experiment costs \$8.07 at public list prices.
Proposal validity differed sharply by model: invalid rates 2.4\% (Haiku),
5.1\% (Opus) and \textbf{79.8\%} (Sonnet), the last because only 3.6\% of its responses
closed without truncation before the output cap. Sonnet is therefore treated as a parsing
artifact, leaving a two-tier capability comparison.

\section{Statistical procedures}
\label{app:stats}

All empirical confidence intervals and arm comparisons in this paper use nonparametric
bootstrap procedures.

\textbf{Intervals.} Reported intervals for a mean are percentile-bootstrap intervals. The
resampling unit is a seeded optimizer run, indexed by (optimizer, seed) within a (library,
reference, variant) cell. The point estimate is the ordinary sample mean, and the interval
uses the 2.5th and 97.5th percentiles of bootstrap means. Non-finite values are removed
before resampling and the retained count is reported. Mitigation, curriculum, and recovery
tables use $B=10{,}000$ resamples; the 792-cell allocation sweep uses $B=2{,}000$.

\textbf{Paired comparisons.} Comparisons with a baseline pair runs by optimizer and seed,
which preserves the shared initialization. The statistic is the mean paired difference.
The reported two-sided bootstrap value is
\[
p=2\min\!\left\{\Pr(\bar d^{*}\leq0),\Pr(\bar d^{*}\geq0)\right\},
\]
capped at one. With $B=10{,}000$, the smallest nonzero two-sided value is
$2\times10^{-4}$; a reported $p<10^{-4}$ means that no resampled mean crossed zero and
should be interpreted as a resolution statement rather than an exact tail probability.

\textbf{Multiplicity and cell-level inference.} No familywise-error or false-discovery-rate
correction is applied, so all reported $p$-values are unadjusted. Raw runs are not pooled
across libraries for inference; comparisons are made within their stated cells.

\section{Ordered-assignment sensitivity}
\label{app:ordered}

The primary results use the combination subspace defined in Appendix~\ref{app:coverage}.
Ordered-assignment checks quantify how much the reported values can change when gate
position is also enumerated.

\textbf{Exactly enumerable ordered spaces.} Two ordered spaces were evaluated completely:
\texttt{Eco2C1G3T1}/\texttt{chain\_k4}, with $P(6,4)=360$ designs, and
\texttt{SC1C1G1T1}/\texttt{chain\_k3}, with $P(12,3)=1{,}320$. Table~\ref{tab:ordered-exact}
compares range-normalized regret on the combination and ordered spaces.

\begin{table}[h]
\centering
\small
\caption{Range-normalized regret on the combination subspace and the fully enumerated
ordered-assignment space.}
\label{tab:ordered-exact}
\begin{tabular}{llrlrr}
\toprule
Library & Topology & Ordered designs & Reference & Combination & Ordered \\
\midrule
\texttt{Eco2C1G3T1} & \texttt{chain\_k4} & 360 & L2 & 0.470 & 0.118 \\
\texttt{Eco2C1G3T1} & \texttt{chain\_k4} & 360 & L2\,$d'$ & 0.431 & 0.229 \\
\texttt{Eco2C1G3T1} & \texttt{chain\_k4} & 360 & L3 & 0.547 & 0.436 \\
\texttt{Eco2C1G3T1} & \texttt{chain\_k4} & 360 & L4 & 0.926 & 0.506 \\
\texttt{SC1C1G1T1} & \texttt{chain\_k3} & 1{,}320 & L2 & 0.133 & 0.139 \\
\texttt{SC1C1G1T1} & \texttt{chain\_k3} & 1{,}320 & L2\,$d'$ & 0.230 & 0.302 \\
\texttt{SC1C1G1T1} & \texttt{chain\_k3} & 1{,}320 & L3 & 0.999 & 0.123 \\
\texttt{SC1C1G1T1} & \texttt{chain\_k3} & 1{,}320 & L4 & 0.096 & 0.243 \\
\bottomrule
\end{tabular}
\end{table}

Regret remains positive in all eight ordered cells, but the magnitude changes in both
directions: it is higher in three and lower in five, with mean absolute change 0.273 and
maximum change 0.876. No ordered cell becomes zero-regret. The combination-space values are
therefore neither uniformly conservative nor uniformly anti-conservative.

\textbf{Large ordered spaces.} For the two $P(20,4)=116{,}280$ spaces in
Table~\ref{tab:ordered-est}, the L1 score was evaluated on every ordered design, so the cheap
argmax is exact. The reference maximum was estimated from that design and a uniform sample
of 4{,}000 additional designs. A sampled maximum cannot exceed the true maximum, so the
reported absolute regret is a lower bound on the full ordered-space absolute regret. The
normalized column uses the sampled reference range and is descriptive rather than a formal
lower bound.

\begin{table}[h]
\centering
\small
\caption{Ordered-assignment checks for \texttt{Eco1C1G1T1}, with 116{,}280 designs per
cell. The cheap argmax is exact; the reference statistics use a sample of 4{,}001 designs.}
\label{tab:ordered-est}
\begin{tabular}{llrlr}
\toprule
Library & Topology & Reference & Absolute lower bound [bootstrap] & Normalized \\
\midrule
\texttt{Eco1C1G1T1} & \texttt{chain\_k4} & L2 & 0.164 [0.091, 0.164] & 0.059 \\
\texttt{Eco1C1G1T1} & \texttt{chain\_k4} & L2\,$d'$ & 0.502 [0.258, 0.502] & 0.107 \\
\texttt{Eco1C1G1T1} & \texttt{chain\_k4} & L3 & 0.327 [0.299, 0.327] & 0.257 \\
\texttt{Eco1C1G1T1} & \texttt{chain\_k4} & L4 & 0.100 [0.091, 0.100] & 0.264 \\
\texttt{Eco1C1G1T1} & \texttt{xnor\_k4} & L2 & 0.171 [0.142, 0.171] & 0.066 \\
\texttt{Eco1C1G1T1} & \texttt{xnor\_k4} & L2\,$d'$ & 0.392 [0.303, 0.392] & 0.087 \\
\texttt{Eco1C1G1T1} & \texttt{xnor\_k4} & L3 & 0.331 [0.312, 0.331] & 0.271 \\
\texttt{Eco1C1G1T1} & \texttt{xnor\_k4} & L4 & 1.417 [1.417, 1.417] & 0.974 \\
\bottomrule
\end{tabular}
\end{table}

All eight absolute lower bounds are positive. Validation against six exactly evaluated
ordered cells and 200 replicates per sample size found mean normalized bias of $-0.103$ at
$m=10$, $-0.055$ at $m=50$, $-0.020$ at $m=200$, and $-0.008$ at $m=800$; no replicate
overestimated the exact value. Together, the
exact and sampled checks preserve the qualitative positive-regret finding, but no specific
21{,}362-design value should be transferred to the full ordered space.

\textbf{Subsampling in non-enumerable spaces.} Exhaustive enumeration of the combination
subspace is tractable at $k \le 4$ and becomes prohibitive in larger circuit spaces. Validating
subsampled estimators of Eq.~\ref{eq:regret} against exact values on enumerable cells shows
that uniform-random subsampling of $m$ designs converges toward exact regret with modest
finite-sample bias, whereas restricting a subsample to the top-$m$ designs ranked by cheap
score is biased low throughout the evaluated range. A pre-run screen restricted to
cheap-simulator favorites therefore underestimates exact regret.

\section{Bounds for stronger per-gate verifiers}
\label{app:verifier}

\paragraph{Method.} We evaluate five verifier classes on every (library, topology,
reference) cell. \textbf{V0} is gate-blind. \textbf{V1} blacklists the worst designs by
cheap score. \textbf{V2} imposes a growth floor, and \textbf{V3} a quality floor, using
cheap-simulator features. \textbf{V4} is an oracle whitelist allowed to admit any subset
using full reference scores. V4 is non-implementable and provides an exact upper bound for
the per-gate-admissibility class through subset enumeration.

The regret numerator remains the full combination-subspace optimum, including when the verifier restricts
the admitted set. Each verifier is evaluated at its best parameter subject to admitting at
least 25\% of the design space. This constraint prevents a low regret value obtained by
removing nearly all candidates.

\begin{table}[h]
\centering
\small
\caption{Normalized regret under each verifier class against growth burden (L3), best
parameter subject to admitting $\ge 25\%$ of the space, averaged over topologies. V4 is the
non-implementable oracle whitelist and therefore an upper bound on the whole class.}
\label{tab:verifierladder}
\begin{tabular}{lrrrrr}
\toprule
Library & V0 gate-blind & V1 blacklist & V2 growth floor & V3 quality floor & V4 oracle \\
\midrule
\texttt{Bth1C1G1T1} & 0.346 & 0.150 & 0.150 & 0.273 & 0.035 \\
\texttt{Eco1C1G1T1} & 0.383 & 0.172 & 0.174 & 0.381 & 0.036 \\
\texttt{Eco1C2G2T2} & 0.208 & 0.120 & 0.175 & 0.157 & 0.089 \\
\texttt{Eco2C1G3T1} & 0.636 & 0.314 & 0.314 & 0.636 & 0.314 \\
\texttt{SC1C1G1T1}  & 0.801 & 0.655 & 0.655 & 0.741 & 0.066 \\
\bottomrule
\end{tabular}
\end{table}

\paragraph{Results.} Over all 52 (library, topology, reference) cells at $\ge 25\%$
permissiveness, the gate-blind verifier leaves mean normalized regret 0.344 (max 0.999).
The oracle whitelist reduces this to mean 0.058 and improves on V0 in 45 of 52 cells, with a
median reduction of 0.239. Positive regret remains in 30 cells, with a maximum of 0.511.
On the growth-burden axis, 17 of 20
cells retain nonzero regret under V4 (mean 0.108); the three exceptions are
\texttt{Bth1C1G1T1}/\texttt{chain\_k3}, \texttt{Bth1C1G1T1}/\texttt{chain\_k4} and
\texttt{Eco2C1G3T1}/\texttt{chain\_k3}. On the cytometry axes the bound is tighter: max
0.158 (L2) and 0.203 (L2\,$d'$).

The V4 column bounds the residual for verifiers that admit or reject designs through
per-gate admissibility. This class cannot eliminate fidelity regret across the evaluated
cells. The bound does not apply to a verifier that measures the reference axis or reshapes
the design space.

\paragraph{Cheap features as constraints.} Two cheap features predict simulator
discrepancy in some libraries. We evaluate three uses: a hard constraint, a
lexicographic preference, and a soft penalty against the naive cheap-argmax baseline and
against two paid baselines (Fig.~\ref{fig:featdisc}). At their \emph{best} parameter per cell, the free constraints help
in 30/52 (hard), 32/52 (lexicographic) and 26/52 (soft) cells, with median gains over all
cells of 0.054, 0.061 and 0.003. The paid comparators do better: a fitted discrepancy model
helps in 37/46 cells (median 0.233) and the reference oracle in 46/52 (median 0.293).
Medians are over every cell, including those the method does not help; over helping cells
only they are 0.158, 0.181, 0.152, 0.295 and 0.343.

\begin{figure}[h]
\centering
\includegraphics[width=0.92\textwidth]{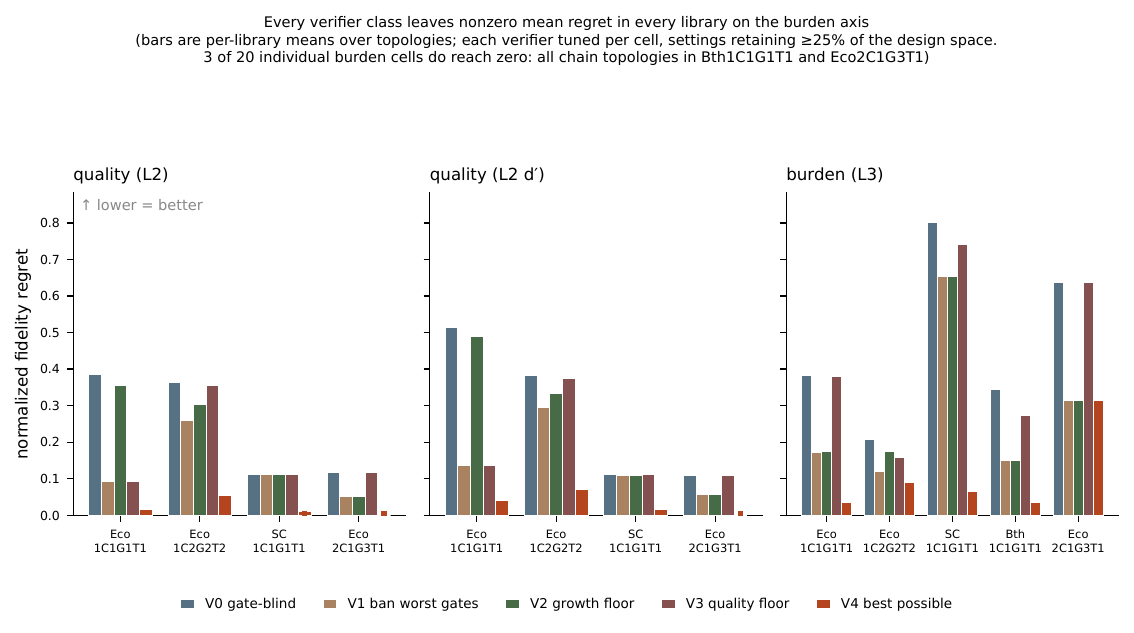}
\caption{Fidelity regret under each verifier class as a function of how much of the design
space the verifier admits. The oracle whitelist (V4) is a non-implementable upper bound on
the per-gate-admissibility class; it flattens well above zero in most cells.}
\label{fig:verifierladder}
\end{figure}

\begin{figure}[h]
\centering
\includegraphics[width=0.92\textwidth]{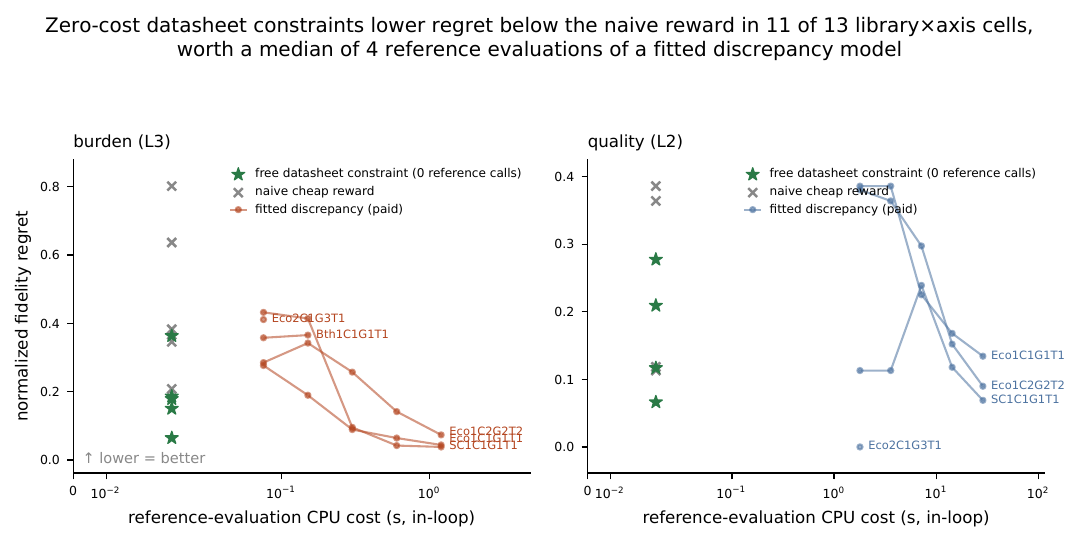}
\caption{Cheap discrepancy-predicting features used as free constraints, against paid
baselines. The best free method beats the naive baseline in 11 of 13 library-by-axis cells,
at a median of 4.0 reference calls to match.}
\label{fig:featdisc}
\end{figure}

\paragraph{Tuning dependence.} The gains above use the best parameter per cell, selected
with reference evaluations. Two checks quantify this dependence. First, at one fixed parameter across all cells, the free constraints improve regret in 22 of 39
(library, reference, method) rows and \emph{worsen} it in the other 17; per method the split
is 9/4 (hard constraint at 0.90), 6/7 (lexicographic at 0.05) and 7/6 (soft penalty at 16).
The fixed parameters improve 22 rows and worsen 17. Second, a threshold tuned on one
reference axis and applied to another carries a penalty in 7 of 12 transfers, up to 0.270
normalized regret; the worst case is
\texttt{Eco2C1G3T1} tuned on L3 and applied to L2\,$d'$.

Matched against paid alternatives at equal budget, the best free method beats the naive baseline in 11
of 13 cells, and the reference budget needed to match the best free method has median 4.0
reference calls (range 1.0--64.0). The observed improvement is comparable to a small number
of reference calls and depends on reference-based tuning.

\paragraph{Limitations.} V4 is an upper bound over subsets of an \emph{enumerated} space; it
does not bound verifiers that reshape the design space itself. The 25\% permissiveness floor
is a choice: at lower floors V4's numbers fall further, but the admitted set stops being a
usable design space. The ``two features predict discrepancy'' premise is library-specific:
it was established on the cytometry-bearing libraries and does not hold uniformly across the
five. Finally, the free-versus-paid comparison equalises reference calls, not wall-clock; a
fitted discrepancy model also costs fitting time that we do not charge it.

\section{Adaptive allocation and an in-flight sentinel}
\label{app:allocation}

\paragraph{Method.} We compare 16 allocation policies at matched budget on every (library,
topology, reference, budget) cell. The adaptive methods are MF-GP-UCB
\citep{kandasamy2016}, BOCA \citep{kandasamy2017}, an approximate implementation of rMFBO
\citep{mikkola2023}, and multi-fidelity best-arm identification \citep{poiani2022}. The
comparators include successive halving \citep{jamieson2016}, Hyperband \citep{li2018},
three $\epsilon$-interleaving controls, the discrepancy-learning arms evaluated in
Appendix~\ref{app:mit}, and the two degenerate anchors \texttt{cheap\_only} and
\texttt{reference\_only}.

\paragraph{Results.} \texttt{cheap\_only} incurs mean regret fraction 1.163, while the
other policies range from 0.030 to 0.557. The assumption-free successive-halving
floor (both elimination rates) is significantly better than cheap-only in 990 of 1152 cells
(85.9\%); the four adaptive multi-fidelity methods are better in 1847 of 2304 cells
(80.2\%).
At the budgets where the comparison is meaningful ($\ge 128$ reference calls), the best
adaptive policy differs from the best non-adaptive control by at most 0.106 normalized regret
across all nine (library, reference) pairs, and by 0.000 in five. At these budgets, the
best simple policy is one of
\texttt{reference\_only}, \texttt{cheap\_then\_reference} or a discrepancy arm.

\begin{figure}[h]
\centering
\includegraphics[width=0.92\textwidth]{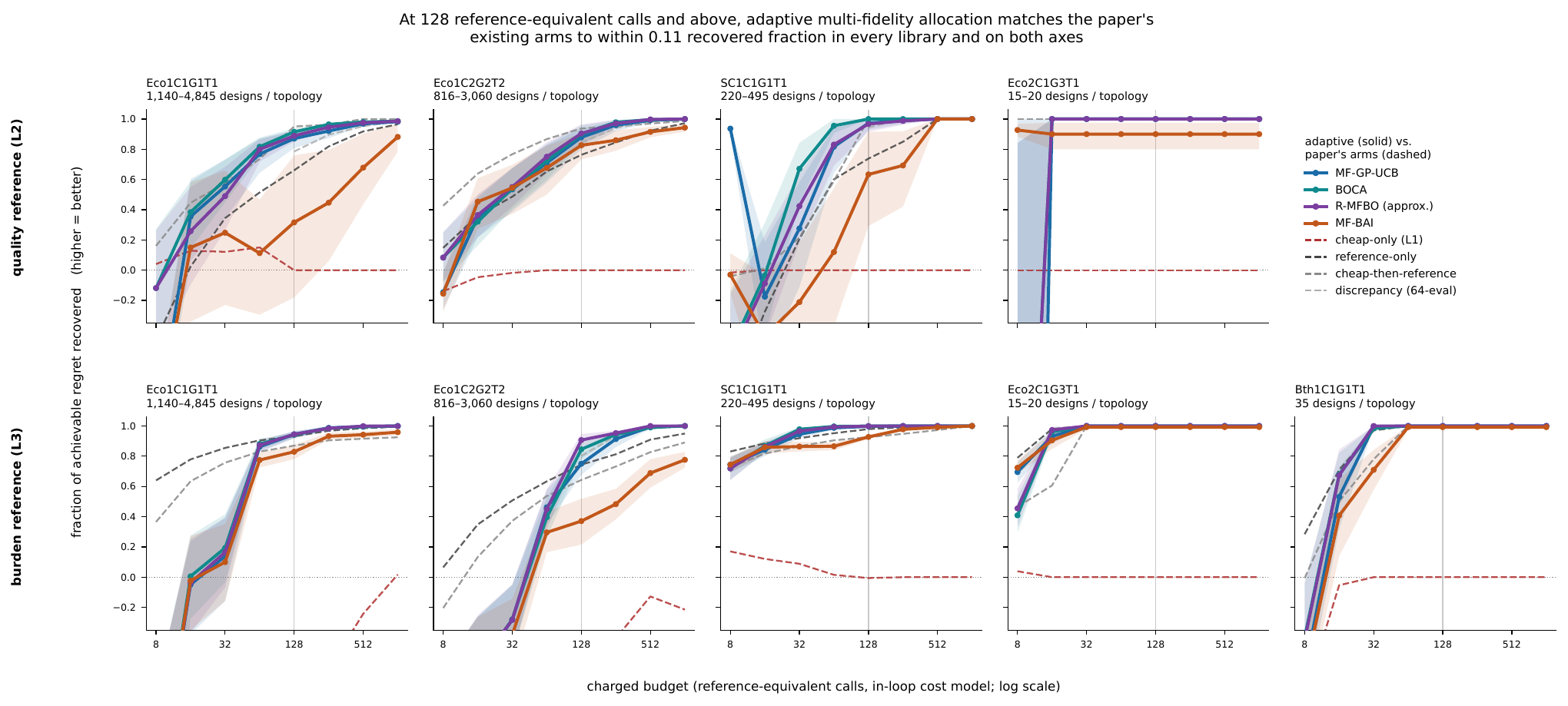}
\caption{Regret against budget for each allocation policy. The adaptive multi-fidelity
methods and the best simple policies have similar frontiers.}
\label{fig:allocfrontier}
\end{figure}

\paragraph{Inferred bias and enumerated regret.} Four policies estimate the cheap
simulator's fidelity bias during optimization. Across two reference axes, the largest
correlation between this estimate and enumerated fidelity regret is
$|\rho_s|=0.293$, with minimum $p=0.271$; all four quality-axis correlations are negative.

With 16 cells on the quality axis
and 20 on the burden axis, the design has 80\% power only for $|r| \ge 0.651$ and
$|r| \ge 0.591$ respectively; the largest correlation observed anywhere is 0.293. The
strongest single case (MF-GP-UCB, quality axis, $\rho_s = -0.293$, $n = 16$) has a
95\% Fisher interval of $[-0.689,+0.237]$, which includes moderate correlations of either
sign. The data exclude a strong observed association at this scale but do not establish
independence. Consequently, the inferred bias estimate does not provide an empirically
validated replacement for the pre-run screen.

\begin{figure}[h]
\centering
\includegraphics[width=0.92\textwidth]{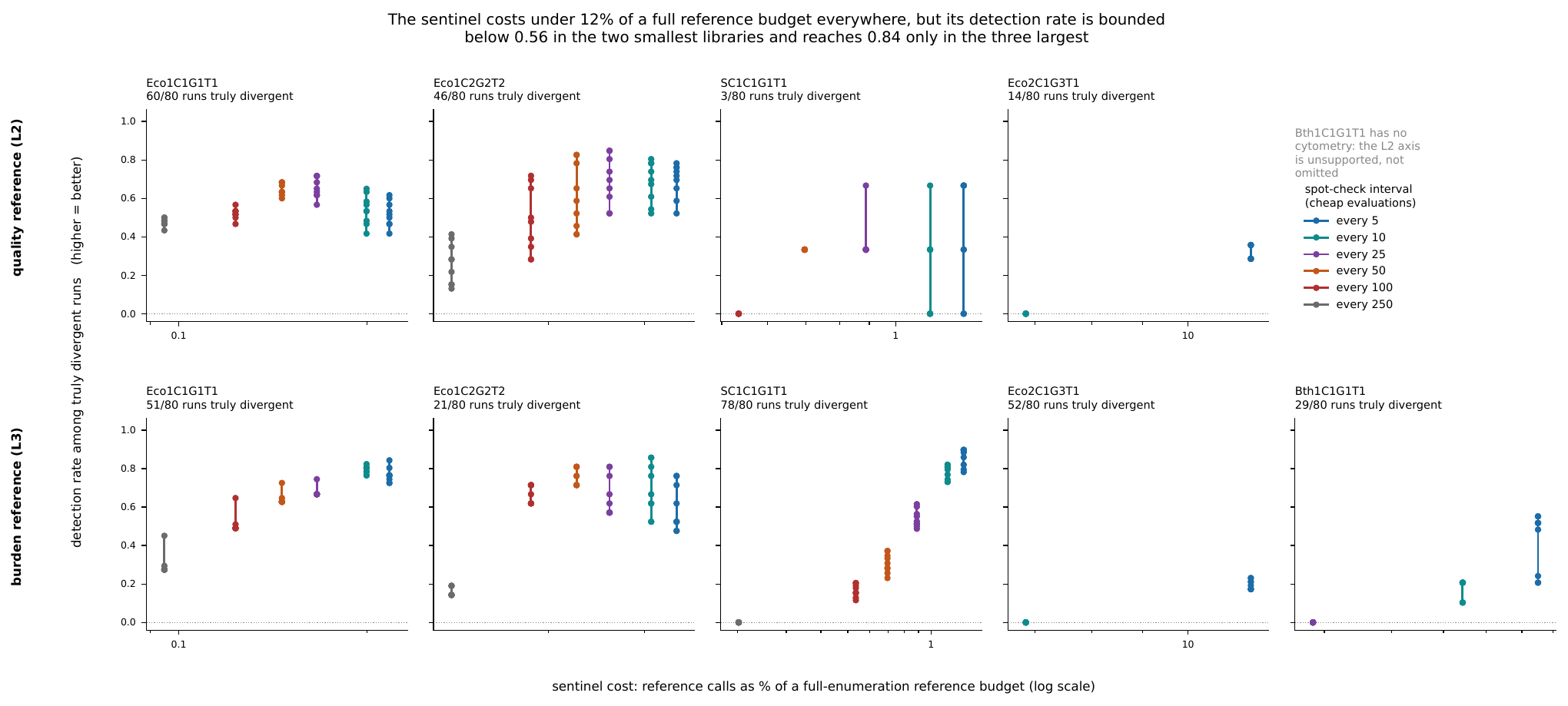}
\caption{Sentinel detection rate against cost, as a fraction of a full reference sweep.}
\label{fig:sentinel}
\end{figure}

\paragraph{In-flight sentinel.} We evaluate a sentinel that spends a small fraction of the budget
re-scoring a sample of the optimizer's current top-$m$ designs at the reference level and
raises an alarm when rank agreement falls below $\tau$
(Fig.~\ref{fig:sentinel}). At the best
operating point of each of the nine (library, reference) cells, detection rate ranges from
0.143 to 0.683 at false-alarm rates at or below 0.200, for a cost between 0.09\% and 11.57\%
of a full reference sweep.

Converting detection into recovered regret gives a positive net gain in 7 of 9 cells, up to
$+0.407$, and a negative gain in 2: $-0.056$ for \texttt{Bth1C1G1T1}/L3 and $-0.022$ for
\texttt{SC1C1G1T1}/L2. In these two cells, allocating the sentinel calls to optimization
would yield lower regret.

\paragraph{Cost-model sensitivity.} Sweeping the price of a reference call
gives a median rank Spearman correlation of 0.996
(minimum 0.732) with the baseline ordering. The top policy changes in 19 of 272 sweep
points (7.0\%), so broad rankings are stable while the optimum remains price-dependent.

\paragraph{Limitations.} The four adaptive methods are re-implementations against their
published descriptions rather than the authors' code, and the rMFBO
comparator is approximate. Negative results for these implementations are therefore weaker
evidence than results obtained with the authors' implementations. The design space here is
the enumerated space, which is smaller and more structured than the spaces these methods
target. Budgets are in reference-call equivalents under a modelled cost, not measured
wall-clock. The sentinel's alarm threshold is tuned per cell on the same data it is
evaluated on, so its detection rates are optimistic.

\section{A decision procedure for practitioners}
\label{app:decision}

\paragraph{Method.} We define two decision rules for a cheap simulator, an available
reference assay, and a fixed budget, and validate them on all 360 (library, topology,
reference, budget) cells.
\textbf{Rule-T} spends a fixed fraction $\alpha = 0.34$ of the budget on reference
evaluations of the cheap simulator's top designs, then re-ranks. \textbf{Rule-S} first
spends a diagnostic sample to estimate the cheap--reference rank correlation $\rho$ on the
target reference axis, then either re-ranks, uses the reference alone, or
accepts the cheap pick. Both are compared against always-cheap, always-reference, and two
leave-one-out oracles that bound what any selection rule could achieve.

\paragraph{Results.} Mean excess regret over the per-cell best achievable is 0.016 for
Rule-T and 0.017 for Rule-S, against 0.305 for always-cheap and 0.034 for always-reference.
The leave-one-out selection rules reach 0.014 (leave-one-library-out) and 0.010
(leave-one-axis-out), placing Rule-T within 0.002--0.006 of these reference-informed
comparators. Rule-T beats always-cheap in 330 of 360 cells; against always-reference it wins
in 161 and ties in 37, for 198 cells with equal or lower regret.

Performance varies by axis. Mean excess regret for Rule-T is 0.004 (L2), 0.002
(L2\,$d'$), 0.014 (L3), and 0.038 (L4), compared with 0.170, 0.189, 0.447, and 0.363 for
always-cheap. The largest absolute reduction occurs on the burden axis.

\begin{figure}[h]
\centering
\includegraphics[width=0.92\textwidth]{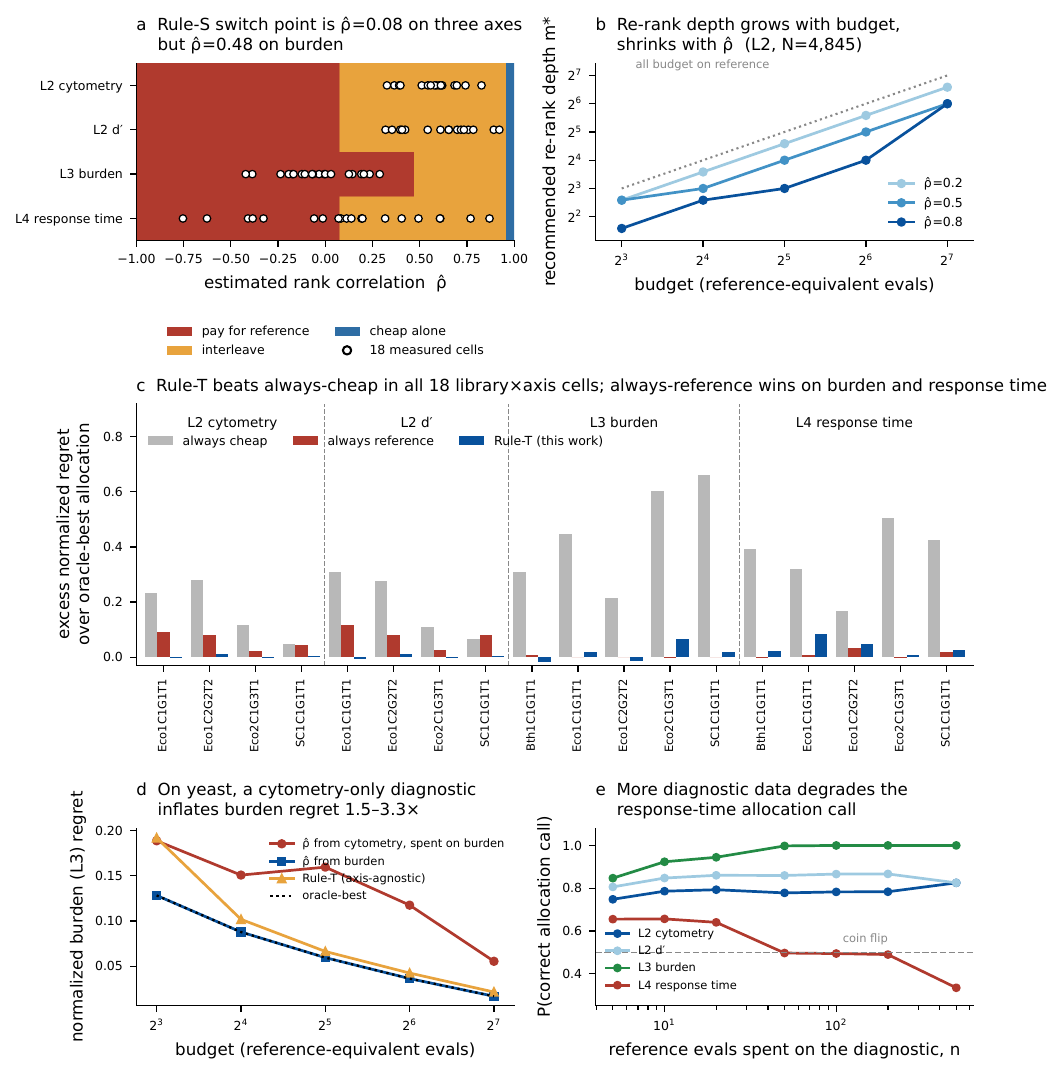}
\caption{The decision procedure and its validation against the enumeration.}
\label{fig:decision}
\end{figure}

\paragraph{Axis transfer.} Rule-S must estimate correlation on the target reference axis.
Running the diagnostic on quality and acting
on burden costs up to 0.355 normalized regret and is worse than diagnosing correctly in 35
of 80 cells. The worst cases are exactly where the
two axes disagree: \texttt{Eco2C1G3T1}/\texttt{and\_k3} has $\rho_{L2} = +0.696$ against
$\rho_{L3} = -0.386$, and \texttt{SC1C1G1T1}/\texttt{and\_k3} has $\rho_{L2} = +0.742$
against $\rho_{L3} = -0.000$. Rule-T requires no diagnostic and beats naive axis transfer
in 45 of 80 rows; it is worse in the remaining 35 by up to 0.297. Rule-T is therefore the
preferred alternative when a diagnostic can only be collected on a different axis.

\begin{figure}[h]
\centering
\includegraphics[width=0.92\textwidth]{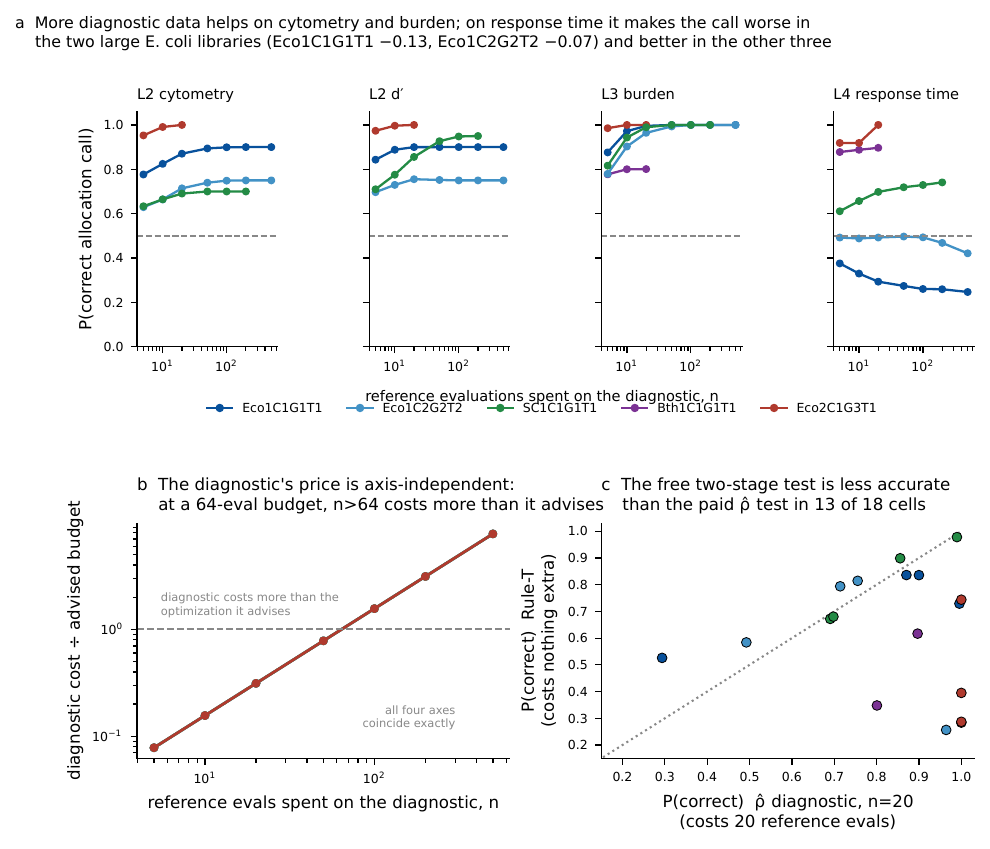}
\caption{Cost of the Rule-S diagnostic as a fraction of the advised budget, and the cells
where it converges on the wrong action.}
\label{fig:diagcost}
\end{figure}

\paragraph{Diagnostic cost and convergence.} The diagnostic sample costs a median 0.78\%
of the advised budget and up to
62.5$\times$ it in the smallest-budget cells (Fig.~\ref{fig:diagcost}). More seriously, the diagnostic is \emph{asymptotically wrong},
converging to the wrong action as the sample grows, in 63 of 360 cells, and its accuracy
degrades with more samples in 59. These results support Rule-T as the default; Rule-S
requires an affordable diagnostic on the target axis.

\paragraph{Limitations.} Both rules are validated on the enumeration used to define regret,
and no additional library provides an external test. The leave-one-out comparators measure
transfer within the available libraries and axes. The value $\alpha=0.34$ was selected on
these data. Of Rule-T's 50 cells above 0.05 excess regret, 26 are on L4 and 14 on L3. Both
rules also require access to a reference assay.

\section{Risk-averse objectives for the cheap simulator}
\label{app:robust}

\paragraph{Method.} We test whether risk-averse functionals of a cheap-simulator ensemble
recover the reference optimum. The objectives are CVaR \citep{rockafellar2002},
mean--variance, and a distributionally robust objective over a Wasserstein ball
\citep{esfahani2018}, at a
sweep of risk parameters, on every cell. The Wasserstein inner problem is solved
exactly through its dual, and the solver's limiting cases are
verified against the point estimate and the worst case.

\paragraph{Results.} Across the 608 rows in which a genuine risk objective is applied
(excluding the point estimate and ensemble-mean baselines), 61 improve regret, 137 worsen
it, and 410 leave it unchanged. The best change is $+0.470$ and the worst is $-0.732$. In
the largest library, \texttt{Eco1C1G1T1}, all 152 objective--parameter rows produce a change
of 0.000 on both axes because the ensemble members agree on the selected optimum.

\begin{figure}[h]
\centering
\includegraphics[width=0.92\textwidth]{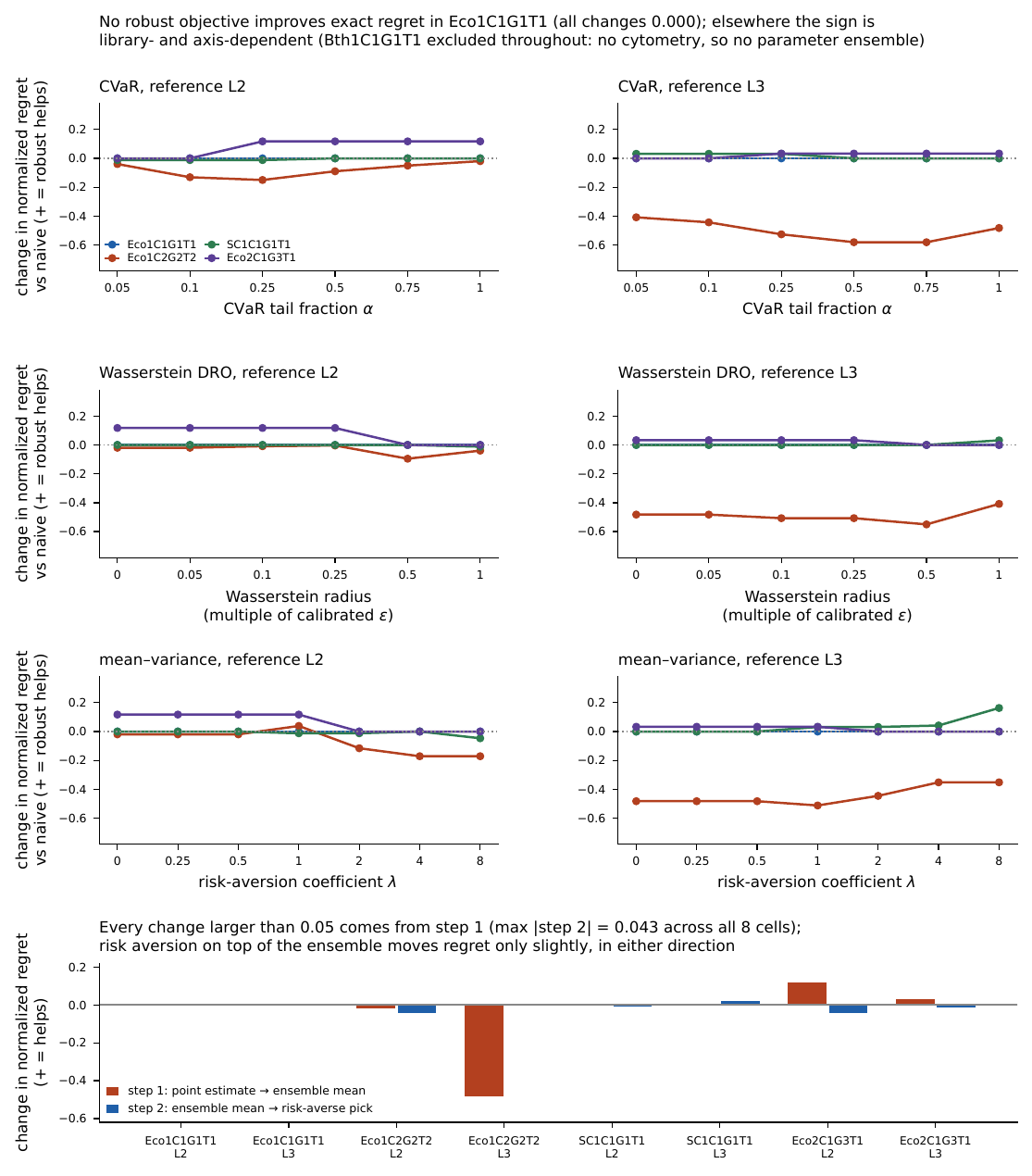}
\caption{Change in normalized regret from each robust objective, relative to the naive
point-estimate pick.}
\label{fig:robust}
\end{figure}

\paragraph{Ensemble and risk components.} Cheap-ensemble risk measures underperform the
naive reward in 3 of 4 training cells. In \texttt{SC1C1G1T1}\,$\times$\,L3, an 8-member
CVaR$_{0.05}$ acquisition improves recovery from 0.007 to 0.152. Decomposing each selection
into two steps
(step 1, replacing the point estimate by the ensemble mean; step 2, moving from the
ensemble mean to the risk-averse pick) shows the change is almost entirely step 1. At the
(library, reference) aggregation, the largest mean step-2 magnitude is 0.043, compared with
a largest mean step-1 magnitude of 0.482. Thus the ensemble substitution accounts for the
gain in this training cell.

At finer granularity, step 1 is larger in 153 of the 178 rows with
$|\Delta|>0.05$ (86.0\%); mean absolute changes are 0.300 for step 1 and 0.073 for step 2.
Across all 608 risk rows, step 2 reaches an absolute change of 0.691, and 86 rows exceed
0.05 from step 2 alone. Of the three (library, reference) cells with a material mean change,
\texttt{Eco1C2G2T2}/L2 is step-2 dominated ($-0.041$ compared with $-0.019$). The ensemble
substitution accounts for most of the aggregate effect, while risk attitude can remain
material in individual cells.

\paragraph{Mechanism.} Modelled spread has weak association with reference shortfall: the
median $|$Spearman(ensemble SD, reference
shortfall)$|$ over 28 cells is 0.094, and only 4 of 28 exceed 0.3. This weak association limits the ability of
risk-averse objectives to identify reference failures.

\paragraph{Limitations.} The ensemble is built by perturbing the cheap simulator's
parameters, so it represents parametric uncertainty. Structural misspecification is absent
from this uncertainty model, which may explain the uninformative spread. Ensembles that
represent structural uncertainty may behave differently. The risk
parameter sweep is finite. \texttt{Bth1C1G1T1} is excluded from this analysis because it lacks
the cytometry axes the ensemble is built on.

\section{Relationship to multi-fidelity optimization and reward misspecification}
\label{app:related}

Multi-fidelity optimization uses correlated low- and high-fidelity evaluations to reduce
the cost of finding a high-fidelity optimum
\citep{kennedy2001,forrester2007,kandasamy2016,kandasamy2017}. Fidelity
regret evaluates a prerequisite for that strategy: whether optimizing the low-fidelity
source selects a high-quality design under the reference. Exact combination-subspace enumeration measures this
relationship before optimizer behavior is introduced. Appendix~\ref{app:allocation} then
compares adaptive multi-fidelity methods with assumption-light allocation rules on the same
spaces.

Lee et al. study screening regret by ranking candidates with a low-fidelity score, evaluating
the top candidates at high fidelity, and reporting their shortfall as a diagnostic for
multi-fidelity Bayesian optimization on molecular and materials benchmarks
\citep{lee2026}. Their paper presents this diagnostic without a formal definition or range
normalization. Fidelity regret belongs to the same family, and range normalization is our
reporting convention. Rule-T is a budgeted form of their surrogate-free screening baseline;
our contribution is its exact validation against the enumeration and the axis-dependence
finding.

Bandit and best-arm-identification methods \citep{jamieson2016,li2018,poiani2022} motivate the successive-halving and direct
reference-allocation comparators. They treat high-fidelity calls as a limited sampling
resource and require fewer structural assumptions about the discrepancy. The decision rule
in Appendix~\ref{app:decision} follows this perspective by reserving a fixed budget fraction
for reference evaluation of high-ranked cheap candidates.

Distributionally robust optimization motivates the CVaR, mean--variance, and
Wasserstein-ball objectives \citep{rockafellar2002,esfahani2018} in
Appendix~\ref{app:robust}. These objectives address
uncertainty represented by the cheap-simulator ensemble. Their effectiveness depends on
ensemble spread predicting reference shortfall, which is weak in the evaluated cells.

Reward hacking and Goodhart-style proxy optimization describe failures caused by optimizing
a proxy that diverges from the intended objective
\citep{amodei2016,skalse2022,manheim2018}. This paper measures the established
proxy--reference shortfall exactly on an enumerable scientific-design space. Its distinction
from learned reward-model overoptimization \citep{gao2023} is that
the misspecification is fixed by the simulator pair and can be measured before policy
training.

\end{document}